\documentclass[preprint,12pt]{elsarticle}
\usepackage{float}
\usepackage{placeins}
\usepackage{algorithm,algcompatible,amsmath,amssymb,algpseudocode}
\usepackage{multirow}
\usepackage{xcolor}
\usepackage{soul}
\sethlcolor{yellow}
\usepackage{pifont}
\newcommand{\cmark}{\ding{51}}   % ✓
\newcommand{\xmark}{\ding{55}}   % ✗
\usepackage{longtable}
\usepackage{makecell}
\usepackage{multicol}
\usepackage{textcomp}
\usepackage{threeparttable}
\usepackage{rotating}
\usepackage{url}
\usepackage{adjustbox}
\usepackage{colortbl}
\definecolor{lightgray}{gray}{0.92}
\usepackage{caption}
\usepackage{graphicx} 
\usepackage{array}
\newcolumntype{Y}{>{\raggedright\arraybackslash}X}
\usepackage{amsthm}
\usepackage{pifont}
\usepackage{booktabs}
\usepackage{tabularx}
\usepackage[most]{tcolorbox}
\usepackage[hidelinks]{hyperref}
\usepackage[pagewise]{lineno}

\ifdefined\pdfgentounicode
  \input{glyphtounicode}
\fi
\usepackage{cmap}
\usepackage[T1]{fontenc}
\usepackage{lmodern}
\usepackage[utf8]{inputenc}
\usepackage[final]{microtype}
\DisableLigatures{encoding = *, family = *}
\algnewcommand\INPUT{\item[\textbf{Input:}]}%
\algnewcommand\OUTPUT{\item[\textbf{Output:}]}%

\begin{document}
\begin{frontmatter}



\title{Multi-Task Crack Foundation Model for Engineering-Reliable Crack Representation and Topology Preservation in Civil Infrastructure}

\cortext[cor1]{Corresponding author}

\author[label1]{Blessing Agyei Kyem}
\ead{blessing.agyeikyem@ndsu.edu}
\author[label1]{Joshua Kofi Asamoah}
\ead{joshua.asamoah@ndsu.edu}
\author[label1]{Eugene Denteh}
\ead{eugene.denteh@ndsu.edu}
\author[label1]{Armstrong Aboah\corref{cor1}}
\ead{armstrong.aboah@ndsu.edu}

\affiliation[label1]{
    organization = {Department of Civil, Construction and Environmental Engineering, North Dakota State University}, 
    city         = {Fargo},
    postcode     = {58102}, 
    state        = {ND},
    country      = {USA}
}
% \affiliation[label2]{
%     organization = {Department of Civil, Construction and Environmental Engineering, University of Memphis}, 
%     city         = {Memphis},
%     postcode     = {38152}, 
%     state        = {TN},
%     country      = {USA}
% }


% \author{} %% Author name

% \affiliation{organization={},%Department and Organization

%% Abstract
\begin{abstract}
Reliable crack assessment requires not only accurate pixel-level masks but also connected crack geometry and confidence estimates that remain stable under domain shift. However, existing segmentation models can achieve high overlap scores while fragmenting cracks, missing fine branches, and providing no calibrated uncertainty. To address this gap, this paper proposes CrackGeoFM, a multi-task framework that combines a frozen visual foundation backbone with crack-specific adaptation for mask prediction, skeleton reconstruction, and uncertainty estimation.  The framework integrates a Frequency-Guided Crack Enhancement Module (FCEM) to enhance high-frequency crack cues, a Crack-Domain Feature Adaptation Module (CFAM) to adapt frozen backbone features to crack-domain patterns, and a Structure-Aware Multi-Task Decoder (SMTD) to jointly decode masks, skeletons, and uncertainty. Across 20 crack datasets, CrackGeoFM achieves state-of-the-art segmentation, improved topology preservation, calibrated uncertainty, and effective few-shot adaptation with only five labeled images. These results support reliable, generalizable, and engineering-oriented crack analysis for infrastructure assessment.
\end{abstract}


% \begin{highlights}
% \end{highlights}

%% Keywords

\end{frontmatter}

% \linenumbers

\section{Introduction}
% \linenumbers
\label{intro}
Automated crack analysis has become an important part of image-based civil infrastructure inspection because cracks in pavements, bridges, tunnels, concrete members, masonry walls, and related assets often reflect traffic loading, aging, material fatigue, environmental exposure, or structural distress. Early crack detection therefore enables infrastructure owners to monitor deterioration and plan maintenance before damage becomes more severe \cite{yuan2024review,yang2018fcn,dung2019fcn}. To support this objective, the increasing use of handheld devices, vehicle-mounted cameras, UAVs, and robotic platforms has made it possible to collect inspection images at scale, creating new opportunities for computer vision methods to detect and map cracks automatically \cite{zhang2016roadcrack,liu2019unet,owora2025pavesam}. Within this automated workflow, segmentation is especially useful because it produces pixel-level crack maps that support estimates of crack area, length, connectivity, and spatial distribution, rather than only a crack/non-crack decision \cite{yang2018fcn,ji2020quantification,chen2024multitask}. The need for reliable pixel-level crack maps and large-scale visual data have therefore made deep learning-based crack segmentation a central research direction.

Although deep learning has substantially advanced crack segmentation, reliable crack analysis remains difficult in real inspection settings. Inspection images vary widely in material type, viewpoint, lighting, resolution, crack scale, background texture, and annotation style. Cracks are also thin, sparse, often discontinuous, and visually similar to shadows, stains, joints, rough surface texture, and other non-crack patterns. As a result, models that perform well on standard benchmarks may still degrade when applied to new infrastructure domains or imaging conditions. Public crack datasets further show that segmentation performance remains sensitive to foreground sparsity, limited labels, and domain shift \cite{zou2012cracktree,enhancing,pavesync,zou2019deepcrack,eisenbach2017gaps,task_specific,goo2025hybrid,asadi2026cracksegflow,big_data}. This gap between benchmark accuracy and cross-domain reliability motivates a shift beyond stronger task-specific segmenters toward foundation models and transferable crack representations.

Foundation models are especially relevant to cross-domain crack analysis because large-scale pretraining can produce visual features that transfer across diverse inspection settings \cite{kirillov2023sam,oquab2024dinov2,pavefm}. Recent crack and infrastructure damage studies show that these representations can be adapted for segmentation tasks \cite{ge2024cracksam,owora2025pavesam,ye2024sambased}. However, most foundation-model-based crack methods still focus primarily on producing segmentation masks. This mask-centered formulation is insufficient for engineering-reliable crack analysis because cracks are thin, connected, and often branching structures rather than only foreground regions. A predicted mask may achieve reasonable Dice or IoU while breaking crack continuity, missing branches, merging unrelated dark patterns, or assigning high confidence to ambiguous regions. Therefore, a crack-domain foundation model should learn complementary outputs that describe the same crack evidence at multiple levels: masks for localization, skeletons for geometric continuity, and uncertainty maps for confidence-aware review. Learning these outputs jointly can guide a shared representation during training and reduce the inconsistency that often arises when structure and uncertainty are handled only through separate models or post-processing \cite{saam_reflectnet}.

Motivated by this limitation, this paper proposes CrackGeoFM, a crack-domain foundation model for engineering-reliable crack representation in civil infrastructure inspection. The model builds on a frozen pretrained visual foundation backbone and proposes three crack-specific modules: a Frequency-Guided Crack Enhancement Module (FCEM) that extracts high-frequency crack cues through wavelet decomposition and injects them into the adaptation pathway without disrupting the pretrained patch embedding, a Crack-Domain Feature Adaptation Module (CFAM) that specializes frozen backbone features for crack segmentation through lightweight bottleneck adapters, and a Structure-Aware Multi-Task Decoder (SMTD) that jointly predicts segmentation masks, crack skeletons, and per-pixel uncertainty from multi-scale adapted features. Because the skeleton supervision targets are derived automatically from standard binary ground truth masks, this structural learning requires no additional manual annotation beyond what existing crack datasets already provide. These proposed modules enable CrackGeoFM to produce crack representations that capture not only pixel-level coverage but also the geometric continuity and prediction confidence needed for condition assessment and maintenance decisions.

The contributions of this paper are summarized as follows:
\begin{itemize}
    \item Crack segmentation is formulated as an engineering-reliable 
    representation problem, where the goal extends beyond pixel-level 
    mask accuracy to include structural consistency through skeleton 
    preservation, uncertainty awareness, and cross-domain transfer.
    
    \item CrackGeoFM is proposed as a crack-domain foundation model 
    built on a frozen pretrained visual backbone with three novel 
    modules: a Frequency-Guided Crack Enhancement Module (FCEM) that 
    injects wavelet-derived high-frequency cues into the adaptation 
    pathway, a Crack-Domain Feature Adaptation Module (CFAM) that 
    specializes frozen backbone features for crack segmentation, and 
    a Structure-Aware Multi-Task Decoder (SMTD) that jointly predicts 
    segmentation masks, crack skeletons, and per-pixel uncertainty.
    
    \item A mask-derived topology supervision strategy is introduced 
    in which crack skeletons are generated automatically from ordinary 
    binary masks and used alongside a topology consistency loss to 
    guide structure-aware learning without additional manual labels. Direct skeleton supervision is shown to produce more accurate crack topology than post-hoc skeletonization of mask predictions.
    
    \item A comprehensive evaluation protocol is presented that 
    combines standard segmentation metrics with topology metrics, 
    uncertainty calibration, cross-domain transfer, few-shot 
    adaptation, and efficiency analysis across 20 crack datasets 
    spanning pavements, concrete, masonry, tunnels, and UAV imagery.
\end{itemize}

\section{Related Work} \label{sec:related_work} This section reviews three areas relevant to CrackGeoFM: deep learning-based crack segmentation, foundation models for cross-domain transfer, and topology- and uncertainty-aware crack representation. 

\subsection{Deep Learning-Based Crack Segmentation} 
\label{subsec:deep_learning_crack_segmentation} 
Early image-based crack analysis relied on hand-crafted features to separate cracks from pavement and concrete backgrounds. Zou et al. \cite{zou2012cracktree} proposed CrackTree, which used geodesic shadow removal, crack probability maps, and minimum spanning trees for pavement crack detection. Shi et al. \cite{shi2016cfd} later introduced the CFD dataset and used random structured forests to detect road cracks under challenging surface conditions. These studies showed that crack extraction is difficult because cracks are thin, irregular, discontinuous, and often visually similar to shadows, stains, joints, and surface texture. However, their reliance on hand-designed features limited their robustness across different materials, lighting conditions, and imaging settings. Deep learning addressed this limitation by learning crack features directly from images. Cha et al. \cite{cha2017deeplearning} demonstrated the use of convolutional neural networks for crack damage detection in civil infrastructure images, while Gopalakrishnan et al. \cite{gopalakrishnan2017transfer} showed that transfer learning can improve pavement distress detection when labeled data are limited. Zhang et al. \cite{zhang2016roadcrack} further applied deep CNNs to road crack detection and contributed to large-scale crack analysis through CRACK500. Although these studies improved feature representation, many early CNN-based methods focused on image-level or patch-level detection rather than dense crack segmentation. To produce pixel-level crack maps, later studies adopted fully convolutional and encoder-decoder networks. Fully convolutional networks enabled end-to-end pixel-level prediction for crack segmentation \cite{long2015fcn,yang2018fcn,dung2019fcn}, while U-Net-style encoder-decoder models improved detail preservation through skip connections \cite{ronneberger2015unet,liu2019unet}. Dung and Anh \cite{dung2019fcn} proposed a fully convolutional network for autonomous concrete crack detection, showing that end-to-end segmentation can generate dense crack predictions. Liu et al. \cite{liu2019unet} used U-Net fully convolutional networks for concrete crack detection and showed that skip connections help recover fine crack details lost during downsampling. Dorafshan et al. \cite{dorafshan2018comparison} compared deep convolutional networks with traditional edge detectors and confirmed the advantage of learned features over classical methods. These works established encoder-decoder segmentation as an effective direction for crack analysis, but standard segmentation networks still struggle with thin cracks, fragmented predictions, and severe foreground-background imbalance. Building on these advances, general segmentation architectures have introduced stronger contextual modeling and decoding strategies that are also relevant to crack analysis. DeepLabV3+ improved multi-scale contextual representation through atrous spatial pyramid pooling and encoder-decoder refinement \cite{chen2018deeplabv3plus}. SegFormer introduced efficient Transformer-based representations for semantic segmentation \cite{xie2021segformer}, while Mask2Former unified mask-based decoding across segmentation tasks \cite{cheng2022mask2former}. These models provide stronger general-purpose segmentation backbones and decoders, but their success in crack analysis still depends on how well they preserve thin, sparse, and elongated crack structures. Several crack-specific architectures have therefore introduced multi-scale learning, hierarchical supervision, and feature fusion. Zou et al. \cite{zou2019deepcrack} proposed DeepCrack, which learns hierarchical convolutional features through deeply supervised side outputs. Liu et al. \cite{liu2019deepcrack_neurocomputing} similarly developed a deep hierarchical feature learning architecture that combines low-level spatial details with high-level semantic features. Yang et al. \cite{yang2020fphbn} introduced the Feature Pyramid and Hierarchical Boosting Network to detect pavement cracks with different widths and appearances. Fan et al. \cite{fan2020ensemble} proposed an ensemble of deep convolutional networks for automatic pavement crack detection and measurement. These methods improved crack mask prediction, but their primary objective remained pixel-level foreground-background classification. More recent studies have extended crack segmentation through attention modules, Transformer encoders, hybrid CNN-Transformer designs, self-supervised learning, and state-space models \cite{context_cracknet,chen2020segnet,wang2022transformer,shamsabadi2022vit,li2024hierarchical,goo2025hybrid,kyem2026selfsupervised,song2025mambafuse,cracksegmenter}. Liu et al. \cite{liu2021crackformer} proposed CrackFormer, a Transformer-based network for fine-grained crack detection. Goo et al. \cite{goo2025hybrid} proposed Hybrid-Segmentor, which combines convolutional and Transformer-based representations for fine-grained crack segmentation in civil infrastructure images. Zuo et al. \cite{zuo2024crackmamba} introduced CrackMamba, a topology-aware state-space model for elongated and continuous crack patterns. These studies show that crack segmentation has progressed from basic CNNs to stronger contextual models. Nevertheless, most methods still evaluate success mainly through mask overlap metrics, allowing models to achieve high IoU or Dice scores while producing broken, noisy, or falsely connected crack maps. 

\subsection{Foundation Models and Cross-Domain Transfer for Crack Analysis} \label{subsec:foundation_models_transfer} The limitations of mask-focused segmentation become more serious when models are applied across different infrastructure domains. Crack images vary widely in surface material, camera distance, lighting, viewpoint, resolution, crack scale, background texture, and annotation style, which can cause models trained on one dataset to perform poorly on another. Public datasets such as CrackTree, CFD, CRACK500, DeepCrack, GAPs, CrackLS315, Stone331, and related benchmarks illustrate the diversity of crack imagery and the difficulty of building models that generalize across inspection settings \cite{zou2012cracktree,shi2016cfd,zhang2016roadcrack,zou2019deepcrack,eisenbach2017gaps,goo2025hybrid,asadi2026cracksegflow}. Zhou et al. \cite{zhou2023review} reviewed deep learning-based crack segmentation and showed that performance is strongly affected by dataset type, annotation quality, architecture, and benchmark design. These works show that benchmark accuracy alone does not guarantee cross-domain reliability. Cross-dataset evaluation is therefore essential for assessing inspection reliability. Models often perform well on held-out splits from the same dataset but degrade on different datasets, materials, cameras, or inspection conditions. This suggests that many methods learn dataset-specific patterns rather than general crack representations \cite{pavefm}. To improve generalization, some studies have explored transfer learning and domain adaptation. Gopalakrishnan et al. \cite{gopalakrishnan2017transfer} used pretrained convolutional networks for pavement distress detection and showed that transfer learning can reduce the need for large task-specific datasets. Weng et al. \cite{weng2023uda} investigated unsupervised domain adaptation for crack segmentation to reduce domain gaps between source and target images. Deng et al. \cite{deng2025ucrackda} proposed UCrack-DA, a multi-scale unsupervised domain adaptation method for surface crack segmentation when labeled target-domain images are unavailable. Zhang et al. \cite{zhang2025syntheticuda} used synthetic pavement crack datasets with unsupervised domain adaptation to reduce the gap between generated and real crack images. These studies directly address domain shift, but their primary goal remains improving crack mask prediction rather than learning richer crack representations. Foundation models provide a stronger starting point because they contain broad visual representations learned from large-scale pretraining. Models such as SAM and DINOv2 have shown that large-scale pretraining can produce transferable representations across visual domains \cite{kirillov2023sam,oquab2024dinov2,pavecap}. Recent studies have begun to adapt these models for crack and structural damage segmentation. Ge et al. \cite{ge2024cracksam} fine-tuned a vision foundation model for crack segmentation in civil infrastructure and showed that SAM-based representations can be adapted to crack images. Guo et al. \cite{guo2025samcrack} proposed a Segment Anything Model-based crack segmentation method using low-rank adaptation to reduce the number of trainable parameters. Ye et al. \cite{ye2024samdamage} applied SAM-based instance segmentation to structural damage detection, extending foundation-model segmentation to masonry and structural inspection. Zhou et al. \cite{zhou2025sepsam} proposed SepSAM, a self-evolving prompting strategy that adapts prompts in a data-driven manner for crack segmentation. These studies show that foundation models can improve crack and structural damage segmentation under challenging conditions. However, most existing work still treats the foundation model mainly as a stronger mask generator. This focus improves segmentation accuracy, but it does not fully address the engineering reliability of the predicted crack map. Because cracks are thin, connected, and often branching damage patterns, foundation models for crack analysis should support not only mask prediction but also structure-aware and uncertainty-aware outputs that preserve topology, identify ambiguous regions, and remain reliable under domain shift. \subsection{Topology-, Geometry-, and Uncertainty-Aware Crack Representation} \label{subsec:topology_geometry_uncertainty} Mask-only segmentation is limited because crack maps are often used for engineering measurements such as length, width, area, branching, continuity, and severity. Segment-based pavement crack quantification methods have been used to estimate crack length, width, area, and crack ratio from segmented crack regions \cite{ye2024sambased}. Fan et al. \cite{fan2020ensemble} also combined deep crack segmentation with measurement procedures for automatic pavement crack detection and quantification. These studies show that segmentation masks can support engineering analysis, but they also show that measurements become unreliable when the predicted mask is fragmented, noisy, or falsely connected. To reduce this dependence on post-processing, recent studies have begun to integrate structural information into model learning. Chen et al. \cite{chen2024multitask} proposed a multi-task deep learning framework for crack segmentation and quantification in reinforced concrete structures, jointly learning crack masks and centerline-related information for width estimation. Zuo et al. \cite{zuo2024crackmamba} incorporated topology awareness into crack segmentation through a state-space model designed to capture long and continuous crack morphology. Topology-aware losses have also been used to preserve thin structures more directly. For example, clDice was introduced to improve the centerline consistency of tubular or curvilinear structures \cite{shit2021cldice}, and related topology-aware segmentation strategies have been adapted to crack analysis \cite{li2026toposam}. These studies treat crack structure as part of the learning problem, but most existing methods still focus on one structural cue, such as centerlines or continuity, rather than jointly modeling masks, skeletons, and uncertainty. Uncertainty estimation provides another important dimension of engineering reliability. Real inspection images often contain ambiguous regions caused by shadows, stains, surface roughness, blurred boundaries, lighting variation, and very thin cracks near the imaging limit. Rathnakumar et al. \cite{rathnakumar2023uncertainty} proposed a Bayesian boundary-aware convolutional network to estimate epistemic and aleatoric uncertainty for crack detection. Dos Santos et al. \cite{dossantos2025uncertainty} studied uncertainty quantification for deep learning-based crack segmentation and emphasized that segmentation outputs should be interpreted together with confidence information. These works show that reliable crack analysis requires both accurate localization and an indication of where predictions may need human review. Synthetic-data generation has also been explored as a way to improve robustness under limited labels and domain shift \cite{weather_adaptive}. Han et al. \cite{han2024crackdiffusion} used diffusion-based crack generation to synthesize training data, while Asadi et al. \cite{asadi2026cracksegflow} introduced a crack segmentation framework that includes synthetic-data generation for improving crack analysis across datasets. These approaches can increase data diversity, but they do not by themselves ensure that the final prediction preserves crack topology or communicates uncertainty. 

In summary, prior work has advanced crack segmentation through stronger architectures, transfer learning, domain adaptation, foundation-model adaptation, topology-aware learning, uncertainty estimation, and synthetic data. However, these directions are often studied independently, and most existing methods remain centered on mask prediction. This leaves a gap for a unified crack-domain foundation model that jointly supports pixel-level localization, structural continuity, confidence-aware interpretation, and cross-domain transfer. CrackGeoFM addresses this gap by combining crack-domain adaptation, frequency-guided enhancement, mask-derived skeleton supervision, topology consistency, and uncertainty prediction within a single foundation-model framework.

\section{Methodology}

This section presents the CrackGeoFM framework and its architectural components. The methodology defines the crack segmentation problem, describes the three proposed modules, the Frequency-Guided Crack Enhancement Module (FCEM), the Crack-Domain Feature Adaptation Module (CFAM), and the Structure-Aware Multi-Task Decoder (SMTD), and details the mask-derived skeleton target generation and loss functions. Figure~\ref{fig:overall_framework} presents the overall architecture.

\subsection{Problem Formulation and Overview}

\begin{figure}[H]
    \centering
    \includegraphics[width=1.0\textwidth]{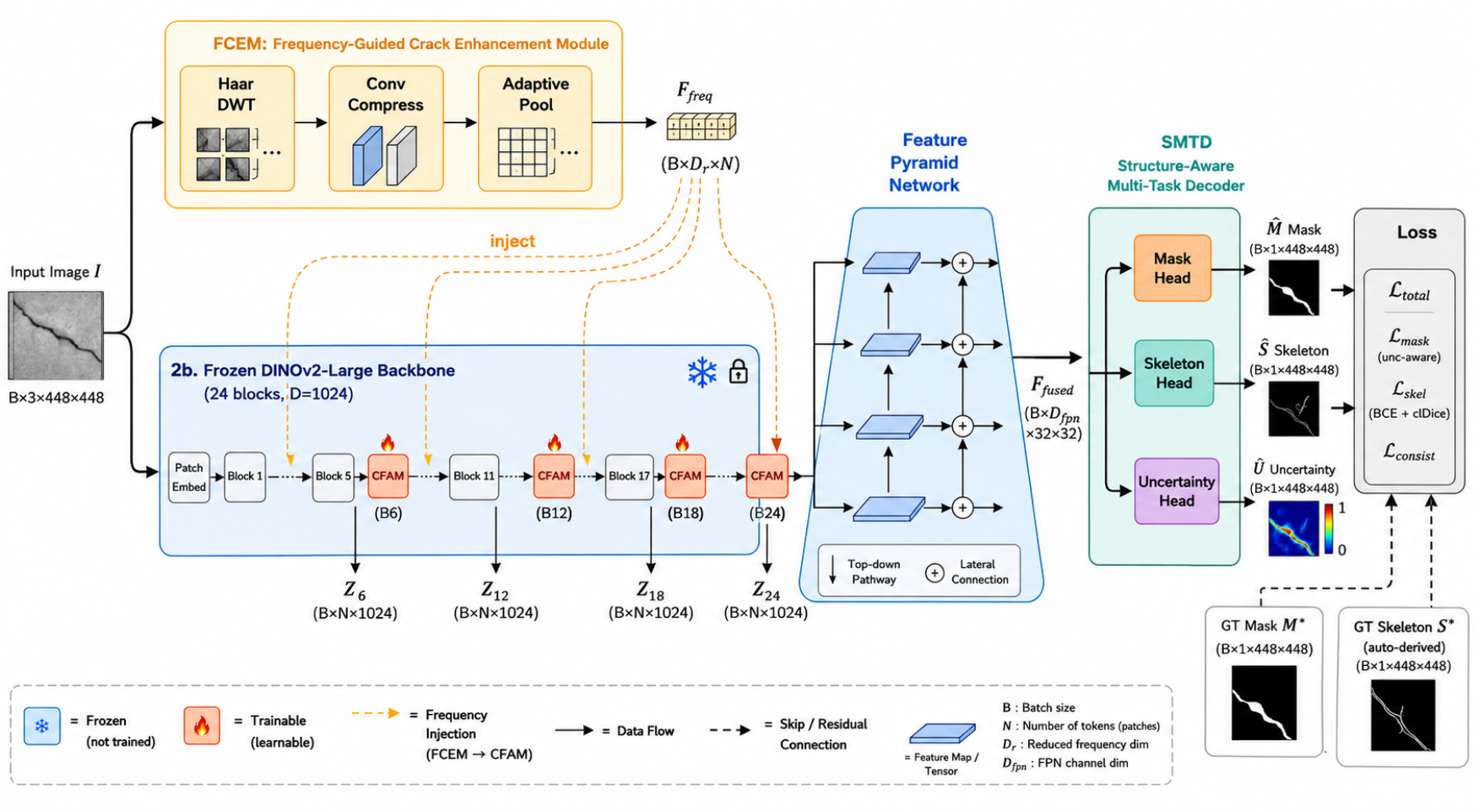}  % change file name and width as needed
    \caption{Overall architecture of CrackGeoFM}
    \label{fig:overall framework}
\end{figure}

Given a batch of input images $I \in \mathbb{R}^{B \times C \times H \times W}$, where $B$, $C$, $H$, and $W$ denote the batch size, channels, height, and width, conventional crack segmentation predicts binary maps $M \in \{0,1\}^{B \times 1 \times H \times W}$, where each pixel is labeled as crack or non-crack. However, a mask alone provides an incomplete description of crack damage: it can achieve reasonable Dice or IoU while breaking continuous cracks, missing branches, merging unrelated dark regions, or confidently labeling shadows as damage. In engineering practice, repair prioritization, growth monitoring, and structural assessment depend on geometric continuity and spatial connectivity, not only pixel coverage. Figure~\ref{fig:overall framework} presents the overall architecture of the proposed framework.

CrackGeoFM reformulates crack segmentation as a multi-output representation problem. The model jointly estimates three complementary outputs from a shared representation: a segmentation mask $\hat{M} \in [0,1]^{B \times 1 \times H \times W}$ that localizes crack regions, a skeleton map $\hat{S} \in [0,1]^{B \times 1 \times H \times W}$ that captures the geometric centerline and connectivity of each crack, and a per-pixel uncertainty map $\hat{U} \in \mathbb{R}^{B \times 1 \times H \times W}$ that quantifies prediction confidence. Formally, the model $\mathcal{F}$ learns:
\begin{equation}
    \mathcal{F}(I; \theta) = \{\hat{M}, \hat{S}, \hat{U}\},
\end{equation}
where $\theta$ denotes the trainable parameters. All supervision targets beyond the binary mask are derived automatically from existing annotations: skeleton targets are generated through morphological skeletonization of ground truth masks, requiring no additional manual labeling.

The framework achieves this objective through a frozen pretrained backbone augmented with three modules. The FCEM extracts high-frequency crack cues through wavelet decomposition and injects them into the adaptation pathway. The CFAM specializes frozen backbone features for crack segmentation through lightweight bottleneck adapters that receive frequency-enhanced information from the FCEM. The SMTD aggregates multi-scale adapted features and produces the three outputs through dedicated prediction heads. The following subsections describe each module in detail.

\subsection{FCEM: Frequency-Guided Crack Enhancement Module}

\begin{figure}[H]
    \centering
    \includegraphics[width=1.0\textwidth]{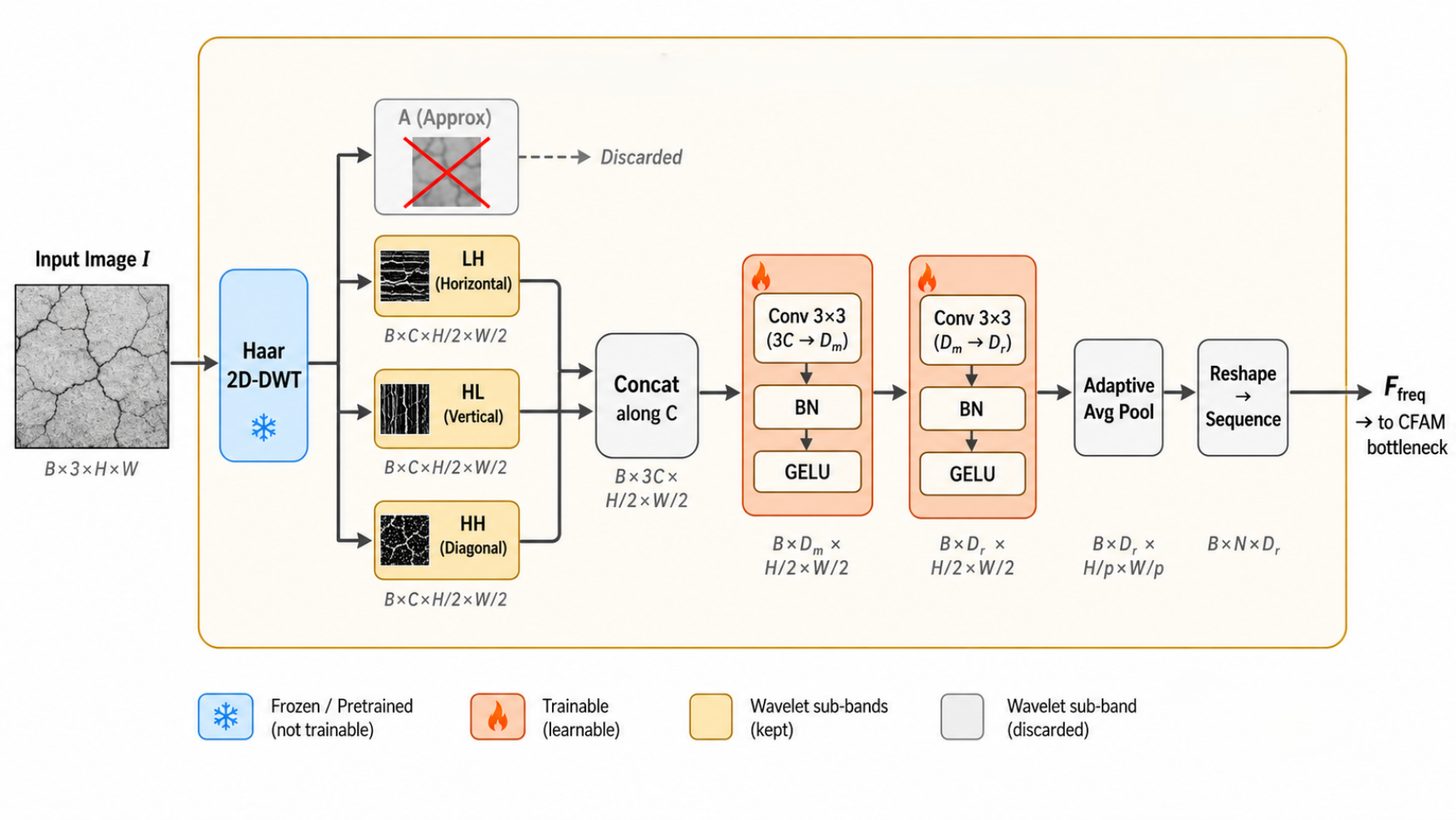}  % change file name and width as needed
    \caption{Architecture of the Frequency-Guided Crack Enhancement 
    Module (FCEM)}
    \label{fig:fcem}
\end{figure}

Cracks are thin, high-frequency structures that can appear faint against textured backgrounds. Standard Vision Transformer patch embeddings linearly project raw patches, which can attenuate fine spatial details that distinguish hairline cracks from surrounding texture. Replacing the pretrained patch embedding would introduce a distribution shift that degrades learned attention patterns. The FCEM addresses this by operating as a parallel side-channel that extracts high-frequency information and injects it into the adaptation pathway, leaving the frozen patch embedding undisturbed. Figure~\ref{fig:fcem} illustrates the internal architecture of the FCEM.

Given an input image $I \in \mathbb{R}^{B \times C \times H \times W}$, the FCEM applies a two-dimensional discrete wavelet transform (DWT) using a Haar basis to each channel independently. For the $c$-th channel $I_c \in \mathbb{R}^{H \times W}$, the DWT decomposes it into a low-frequency approximation sub-band $\mathbf{A}_c$ and three high-frequency detail sub-bands: the horizontal detail $\mathbf{LH}_c$, the vertical detail $\mathbf{HL}_c$, and the diagonal detail $\mathbf{HH}_c$:
\begin{equation}
    \text{DWT}(I_c) = \{\mathbf{A}_c, \mathbf{LH}_c, \mathbf{HL}_c, \mathbf{HH}_c\},
\end{equation}
where each sub-band has spatial dimensions $\frac{H}{2} \times \frac{W}{2}$. Since cracks produce strong responses in the high-frequency detail sub-bands while smooth regions do not, the approximation sub-band $\mathbf{A}_c$ is discarded. The retained detail sub-bands are concatenated across all channels:
\begin{equation}
    F_{\text{freq}} = \text{Concat}\bigl(\mathbf{LH}_1, \mathbf{HL}_1, \mathbf{HH}_1, \ldots, \mathbf{LH}_C, \mathbf{HL}_C, \mathbf{HH}_C\bigr) \in \mathbb{R}^{B \times 3C \times \frac{H}{2} \times \frac{W}{2}}.
\end{equation}

Two successive $3 \times 3$ convolutional layers with batch normalization and GELU activation compress this into a compact representation:
\begin{equation}
    F_{\text{freq}}' = \sigma\bigl(\text{BN}(W_2 \ast \sigma(\text{BN}(W_1 \ast F_{\text{freq}} + b_1)) + b_2)\bigr),
\end{equation}
where $W_1 \in \mathbb{R}^{D_m \times 3C \times 3 \times 3}$ and $W_2 \in \mathbb{R}^{D_r \times D_m \times 3 \times 3}$ are learnable convolutional kernels, $b_1 \in \mathbb{R}^{D_m}$ and $b_2 \in \mathbb{R}^{D_r}$ are bias terms, $D_m$ is an intermediate channel dimension, and $D_r$ is the adapter bottleneck dimension. The result is spatially aligned to the patch token grid through adaptive average pooling:
\begin{equation}
    F_{\text{freq}}'' = \text{AdaptiveAvgPool}\bigl(F_{\text{freq}}',\; \bigl(\tfrac{H}{p}, \tfrac{W}{p}\bigr)\bigr) \in \mathbb{R}^{B \times D_r \times \frac{H}{p} \times \frac{W}{p}},
\end{equation}
where $p$ is the patch size (e.g., $p = 14$ for DINOv2). After reshaping to $F_{\text{freq}}'' \in \mathbb{R}^{B \times N \times D_r}$ with $N = \frac{H}{p} \times \frac{W}{p}$, this representation is passed to each CFAM instance for injection into the adapter bottleneck. The FCEM adds approximately 50K trainable parameters and negligible inference overhead.

\subsection{CFAM: Crack-Domain Feature Adaptation Module}

\begin{figure}[H]
    \centering
    \includegraphics[width=1.0\textwidth]{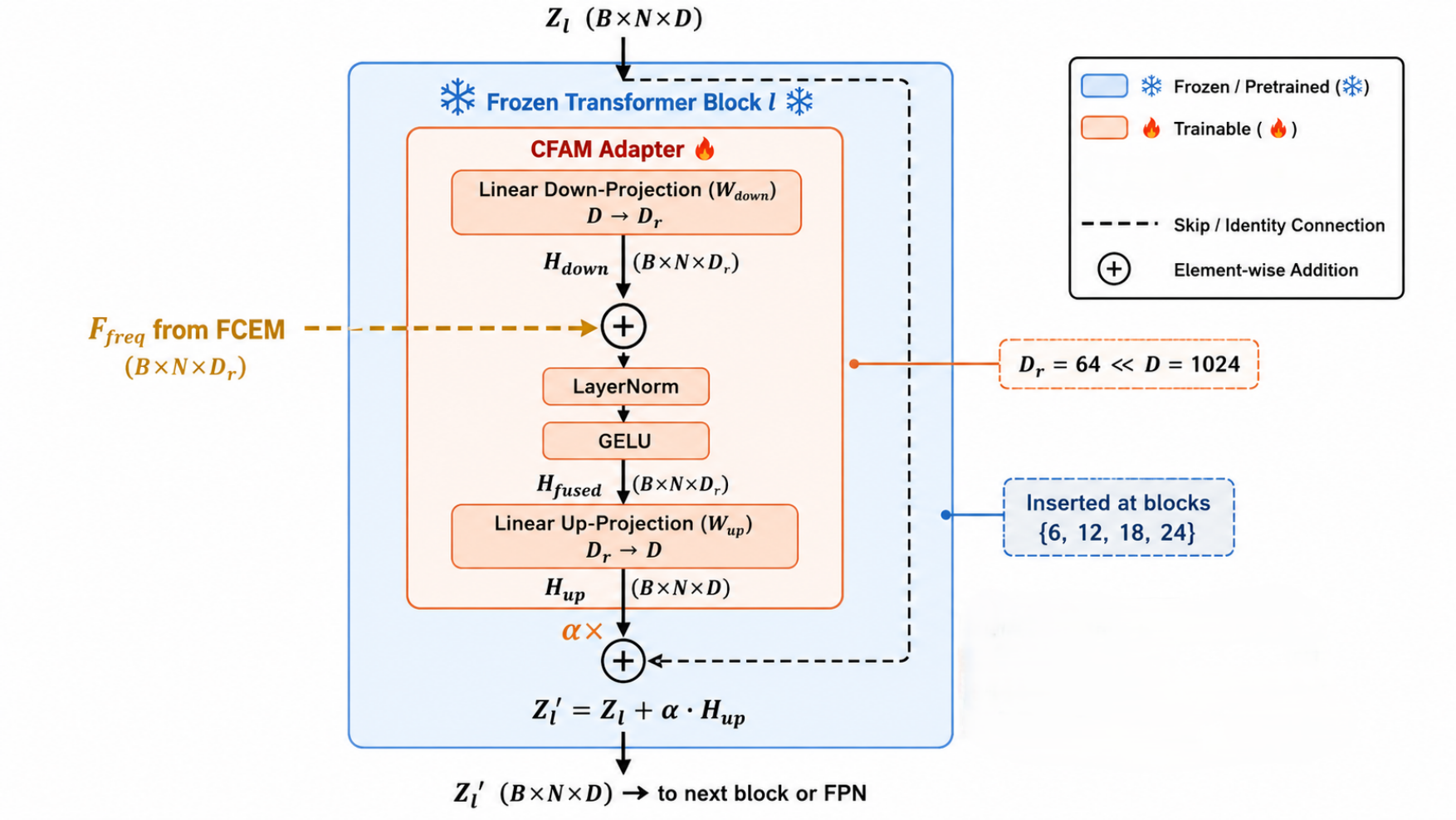}  % change file name and width as needed
    \caption{Architecture of a single Crack-Domain Feature Adaptation Module (CFAM) instance within a frozen transformer block}
    \label{fig:cfam}
\end{figure}

The frozen backbone features are general-purpose visual representations not specialized for crack characteristics such as extreme aspect ratios, sparse foreground, and subtle contrast. The CFAM addresses this by inserting lightweight trainable bottleneck adapters at selected backbone blocks, transforming general features into crack-domain representations while preserving pretrained knowledge. Figure~\ref{fig:cfam} shows the structure of a single CFAM instance within a frozen transformer block.

CrackGeoFM uses a frozen DINOv2-Large backbone with $L = 24$ transformer blocks, hidden dimension $D = 1024$, and 16 attention heads, pretrained on 142 million images~\cite{oquab2024dinov2}. CFAM modules are inserted after blocks $\mathcal{B} = \{6, 12, 18, 24\}$, spanning early, mid, and late representations.

For block $l \in \mathcal{B}$, let $\mathbf{Z}_l \in \mathbb{R}^{B \times N \times D}$ denote the output token sequence. The CFAM first projects tokens to a bottleneck dimension $D_r \ll D$:
\begin{equation}
    \mathbf{H}_l^{\text{down}} = W_l^{\text{down}} \mathbf{Z}_l + b_l^{\text{down}}, \quad \mathbf{H}_l^{\text{down}} \in \mathbb{R}^{B \times N \times D_r},
\end{equation}
where $W_l^{\text{down}} \in \mathbb{R}^{D_r \times D}$ and $b_l^{\text{down}} \in \mathbb{R}^{D_r}$ are learnable parameters. At this bottleneck, the FCEM frequency features are injected through element-wise addition:
\begin{equation}
    \mathbf{H}_l^{\text{fused}} = \sigma\bigl(\text{LN}(\mathbf{H}_l^{\text{down}} + F_{\text{freq}}'')\bigr),
\end{equation}
where $\text{LN}(\cdot)$ is layer normalization and $\sigma$ is GELU. Injecting at the bottleneck allows the adapter to learn joint spatial-frequency patterns rather than treating them as separate streams. The fused representation is projected back to the backbone dimension:
\begin{equation}
    \mathbf{H}_l^{\text{up}} = W_l^{\text{up}} \mathbf{H}_l^{\text{fused}} + b_l^{\text{up}}, \quad \mathbf{H}_l^{\text{up}} \in \mathbb{R}^{B \times N \times D},
\end{equation}
where $W_l^{\text{up}} \in \mathbb{R}^{D \times D_r}$ and $b_l^{\text{up}} \in \mathbb{R}^{D}$. A residual connection preserves the pretrained information:
\begin{equation}
    \mathbf{Z}_l' = \mathbf{Z}_l + \alpha \cdot \mathbf{H}_l^{\text{up}},
\end{equation}
where $\alpha$ is a learnable scaling parameter initialized to 0.1 for training stability.

The four CFAM instances produce adapted feature maps $\{\mathbf{Z}_6', \mathbf{Z}_{12}', \mathbf{Z}_{18}', \mathbf{Z}_{24}'\}$ capturing crack-relevant information at progressively deeper levels: early outputs encode edge and texture patterns, later outputs encode semantic structure. With $D = 1024$ and $D_r = 64$, the total CFAM parameter count is approximately 525K, less than 0.2\% of the frozen backbone.

\subsection{SMTD: Structure-Aware Multi-Task Decoder}

\begin{figure}[H]
    \centering
    \includegraphics[width=1.0\textwidth]{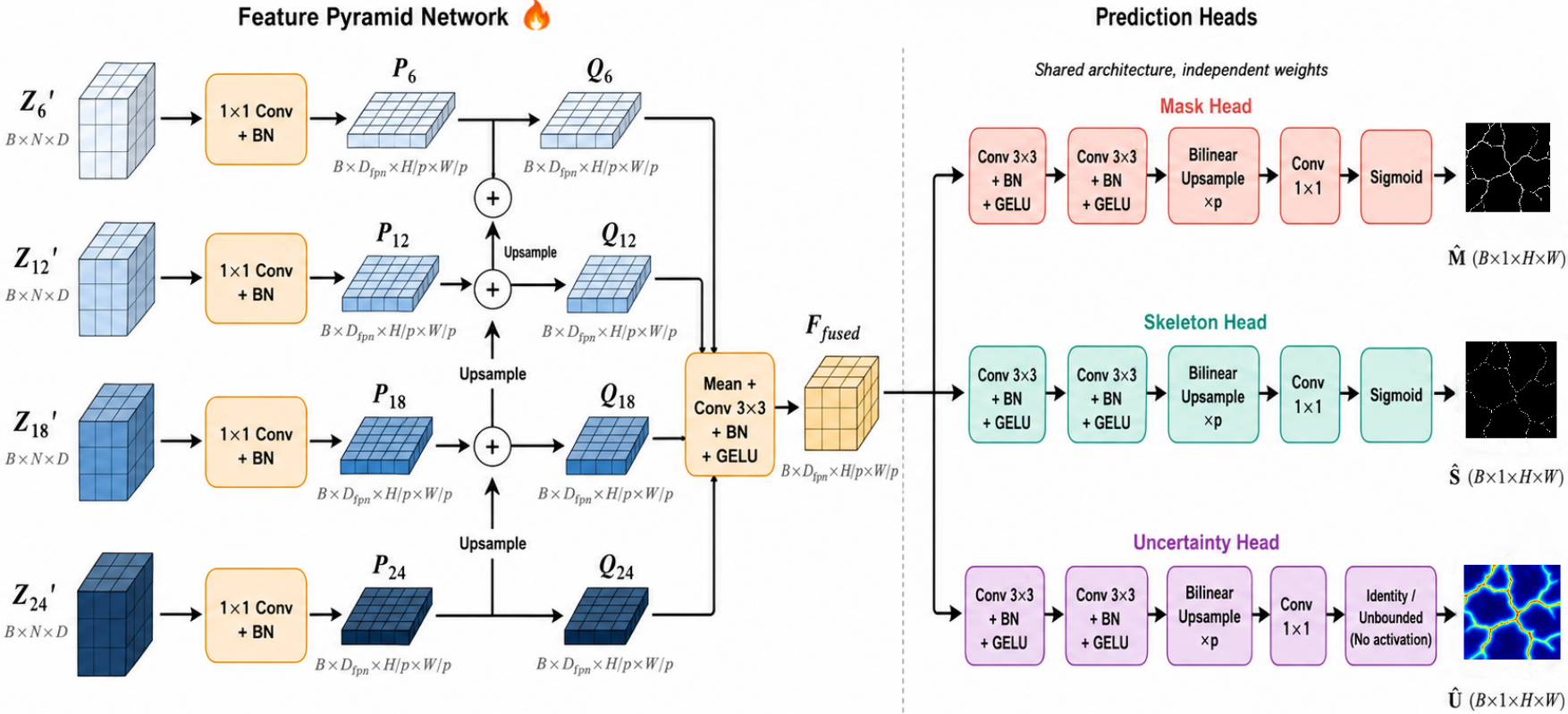}  % change file name and width as needed
    \caption{Architecture of the Structure-Aware Multi-Task Decoder 
    (SMTD).}
    \label{fig:smtd}
\end{figure}

The SMTD aggregates the multi-scale adapted features from CFAM and decodes them into the three target outputs through a feature pyramid and dedicated prediction heads. Figure~\ref{fig:smtd} depicts this two-stage design of the SMTD.

\subsubsection{Multi-Scale Feature Aggregation}

Each adapted output $\mathbf{Z}_l' \in \mathbb{R}^{B \times N \times D}$ is reshaped to a spatial feature map $\mathbf{X}_l \in \mathbb{R}^{B \times D \times \frac{H}{p} \times \frac{W}{p}}$ and projected to a common pyramid dimension $D_{\text{fpn}}$ via a $1 \times 1$ convolution with batch normalization:
\begin{equation}
    \mathbf{P}_l = \text{BN}\bigl(W_l^{1 \times 1} \ast \mathbf{X}_l\bigr), \quad \mathbf{P}_l \in \mathbb{R}^{B \times D_{\text{fpn}} \times \frac{H}{p} \times \frac{W}{p}},
\end{equation}
where $W_l^{1 \times 1} \in \mathbb{R}^{D_{\text{fpn}} \times D \times 1 \times 1}$ is a learnable projection kernel.

A top-down pathway with lateral connections fuses these maps:
\begin{equation}
    \mathbf{Q}_{24} = \mathbf{P}_{24}, \qquad \mathbf{Q}_l = \mathbf{P}_l + \text{Upsample}(\mathbf{Q}_{l+}), \quad l \in \{18, 12, 6\},
\end{equation}
where $l+$ denotes the next deeper level in $\mathcal{B}$ and $\text{Upsample}$ uses bilinear interpolation. The fused representation is obtained by averaging across all levels and applying a smoothing convolution:
\begin{equation}
    \mathbf{F}_{\text{fused}} = \sigma\Bigl(\text{BN}\Bigl(W^{3 \times 3} \ast \frac{1}{|\mathcal{B}|}\sum_{l \in \mathcal{B}} \mathbf{Q}_l\Bigr)\Bigr) \in \mathbb{R}^{B \times D_{\text{fpn}} \times \frac{H}{p} \times \frac{W}{p}},
\end{equation}
where $W^{3 \times 3} \in \mathbb{R}^{D_{\text{fpn}} \times D_{\text{fpn}} \times 3 \times 3}$ is a learnable smoothing kernel and $\sigma$ denotes GELU activation.

\subsubsection{Prediction Heads}

Three independent heads with shared architecture operate on $\mathbf{F}_{\text{fused}}$. Each head $k \in \{\text{mask}, \text{skel}, \text{unc}\}$ consists of a feature refinement block $g_k$ (two successive $3 \times 3$ conv-BN-GELU layers), bilinear upsampling to the input resolution, and a final $1 \times 1$ projection $f_k^{1 \times 1}$ that maps to a single output channel:
\begin{equation}
    \hat{Y}_k = f_k^{1 \times 1}\Bigl(\text{Upsample}\bigl(g_k(\mathbf{F}_{\text{fused}}),\; (H, W)\bigr)\Bigr), \quad \hat{Y}_k \in \mathbb{R}^{B \times 1 \times H \times W}.
\end{equation}

The \textbf{mask head} produces per-pixel crack probabilities via sigmoid activation:
\begin{equation}
    \hat{M} = \sigma_{\text{sig}}\bigl(\hat{Y}_{\text{mask}}\bigr) \in [0,1]^{B \times 1 \times H \times W}.
\end{equation}

The \textbf{skeleton head} predicts the one-pixel-wide crack centerline, encoding topological connectivity, branching structure, and geometric properties critical for crack length measurement and growth monitoring:
\begin{equation}
    \hat{S} = \sigma_{\text{sig}}\bigl(\hat{Y}_{\text{skel}}\bigr) \in [0,1]^{B \times 1 \times H \times W}.
\end{equation}

The \textbf{uncertainty head} estimates per-pixel aleatoric uncertainty as a learned log-variance map, remaining unbounded to represent $\log \sigma_a^2$ where $\sigma_a^2$ denotes the aleatoric variance:
\begin{equation}
    \hat{U} = \hat{Y}_{\text{unc}} \in \mathbb{R}^{B \times 1 \times H \times W}.
\end{equation}
High values of $\hat{U}_{ij}$ indicate ambiguous regions (crack boundaries, shadows, sealed cracks) and serve dual roles: modulating the mask loss during training and flagging low-confidence regions for human review during inference. The multi-task supervision from all three heads produces a richer shared representation than mask-only training, as demonstrated in the ablation study (Section~\ref{sec:ablation}).

\subsection{Mask-Derived Skeleton Target Generation}
\label{sec:topology_targets}
 
\begin{figure}[H]
    \centering
    \includegraphics[width=1.0\textwidth]{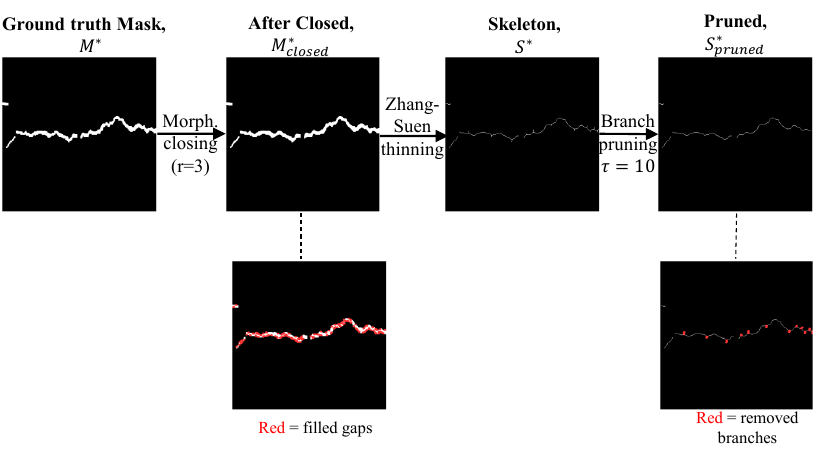}  % change file name and width as needed
    \caption{Mask-derived skeleton target generation pipeline.}
    \label{fig:skeleton_pipeline}
\end{figure}

Skeleton targets are derived automatically from ground truth masks $M^{*} \in \{0,1\}^{H \times W}$ through a three-step morphological pipeline. Figure~\ref{fig:skeleton_pipeline} visualizes the three-step pipeline using a sample ground truth mask from DeepCrack dataset. First, morphological closing with a disk-shaped structuring element $\mathbf{d}_r$ of radius $r = 3$ pixels fills small annotation gaps and connects nearby crack fragments. This radius is large enough to bridge single- and double-pixel gaps introduced by annotation inconsistencies, while remaining small enough to avoid merging spatially distinct cracks or altering the overall crack geometry. Pre-skeleton smoothing through morphological closing is a standard practice for suppressing spurious skeleton branches caused by edge noise in binary masks~\cite{denny2023skeleton, cracktopologyscore2026}:
\begin{equation}
    M^{*}_{\text{closed}} = (M^{*} \oplus \mathbf{d}_r) \ominus \mathbf{d}_r,
\end{equation}
where $\oplus$ and $\ominus$ denote morphological dilation and erosion respectively.
 
Second, the closed mask is reduced to a one-pixel-wide centerline through Zhang-Suen thinning~\cite{zhang1984fast}:
\begin{equation}
    S^{*} = \text{Skeletonize}(M^{*}_{\text{closed}}) \in \{0,1\}^{H \times W}.
\end{equation}

Third, spurious branches shorter than a threshold of $\tau_{\text{prune}} = 10$ pixels are removed to eliminate noise-induced artifacts. At the training resolution of $448 \times 448$, branches below this length correspond to roughly 2\% of the image dimension and are typically caused by jagged annotation boundaries rather than real crack branching. Length-based skeleton pruning is a well-established technique for suppressing parasitic branches in crack skeleton analysis~\cite{denny2023skeleton, li2024automated}:
\begin{equation}
    S^{*}_{\text{pruned}} = \text{Prune}(S^{*}, \tau_{\text{prune}}).
\end{equation}
 
The pruned skeleton serves as the ground truth for the skeleton head. This pipeline introduces zero annotation cost and is executed once as a preprocessing step before training.

\subsection{Loss Formulation}

The total loss combines three components addressing distinct aspects of crack representation. In all loss equations, subscripts $(i,j)$ denote the spatial pixel position across the $H \times W$ image grid.

\subsubsection{Uncertainty-Aware Mask Loss}

The mask head is supervised using binary cross-entropy and Dice loss, modulated by the learned uncertainty. Let $\hat{M}_{ij}$, $M^{*}_{ij}$, and $\hat{U}_{ij}$ denote the predicted mask probability, ground truth label, and predicted log-variance at pixel $(i,j)$ respectively. The uncertainty-aware mask loss is:
\begin{equation}
    \mathcal{L}_{\text{mask}} = \frac{1}{HW} \sum_{i,j} \Bigl[ \frac{1}{2} \exp(-\hat{U}_{ij}) \cdot \ell_{\text{BCE}}(\hat{M}_{ij}, M^{*}_{ij}) + \frac{1}{2} \hat{U}_{ij} \Bigr] + \mathcal{L}_{\text{Dice}}(\hat{M}, M^{*}),
\end{equation}
where $\ell_{\text{BCE}}(\hat{m}, m^{*}) = -[m^{*} \log \hat{m} + (1 - m^{*}) \log(1 - \hat{m})]$ is the pixel-wise binary cross-entropy and the Dice loss is:
\begin{equation}
    \mathcal{L}_{\text{Dice}}(\hat{M}, M^{*}) = 1 - \frac{2 \sum_{i,j} \hat{M}_{ij} \cdot M^{*}_{ij} + \epsilon}{\sum_{i,j} \hat{M}_{ij} + \sum_{i,j} M^{*}_{ij} + \epsilon},
\end{equation}
where $\epsilon = 1 \times 10^{-6}$ is a smoothing constant for numerical stability.
The $\exp(-\hat{U}_{ij})$ term reduces loss contribution from high-uncertainty pixels, while the $\frac{1}{2}\hat{U}_{ij}$ regularizer prevents trivially assigning infinite uncertainty everywhere.

\subsubsection{Skeleton Loss}

The skeleton head uses binary cross-entropy and centerline Dice (clDice)~\cite{shit2021cldice}:
\begin{equation}
    \mathcal{L}_{\text{skel}} = \ell_{\text{BCE}}(\hat{S}, S^{*}_{\text{pruned}}) + \mathcal{L}_{\text{clDice}}(\hat{S}, S^{*}_{\text{pruned}}).
\end{equation}
The clDice loss penalizes topological errors such as broken centerlines and missed branches more heavily than pixel-wise losses, which are dominated by the non-crack background.

\subsubsection{Topology Consistency Loss}

This loss enforces geometric agreement between the mask and skeleton predictions:
\begin{equation}
    \mathcal{L}_{\text{consist}} = 1 - \frac{2 \sum_{i,j} \hat{M}_{ij} \cdot \hat{S}_{ij} + \epsilon}{\sum_{i,j} \hat{S}_{ij} + \sum_{i,j} \hat{M}_{ij} \cdot \mathbf{1}[\hat{S}_{ij} > 0.5] + \epsilon},
\end{equation}
where $\mathbf{1}[\cdot]$ denotes the indicator function that returns 1 when its argument is true and 0 otherwise. This loss penalizes skeleton pixels outside the mask (broken topology) and mask regions disconnected from any skeleton (potential false positives).

\subsubsection{Total Loss}
The combined loss, denoted by \( \mathcal{L}_{\text{total}} \), can be expressed as follows:
\begin{equation}
    \mathcal{L}_{\text{total}} = \mathcal{L}_{\text{mask}} + \lambda_{\text{skel}} \cdot \mathcal{L}_{\text{skel}} + \lambda_{\text{consist}} \cdot \mathcal{L}_{\text{consist}},
\end{equation}
where $\lambda_{\text{skel}}$ and $\lambda_{\text{consist}}$ control the relative importance of each component.

\subsection{Training and Inference}

The model is trained using AdamW with learning rate $\eta = 1 \times 10^{-4}$, weight decay $1 \times 10^{-2}$, cosine annealing with 5-epoch linear warmup, and early stopping (patience 20) over a maximum of 200 epochs with mixed-precision (FP16) training. Since the training set comprises multiple datasets of varying sizes, a temperature-based sampling strategy balances dataset representation:
\begin{equation}
    p_d = \frac{n_d^{1/T}}{\sum_{d'} n_{d'}^{1/T}},
\end{equation}
where $n_d$ is the dataset size and $T = 2.0$ flattens the distribution to give smaller datasets higher relative probability. Augmentation includes random flips, rotation ($\pm 15^{\circ}$), and color jitter, applied consistently to images, masks, and skeleton targets. All images are resized to $448 \times 448$.

During inference, a single forward pass produces all three outputs simultaneously. The mask is thresholded at 0.5 for binary segmentation, the skeleton supports connectivity analysis, and the uncertainty map identifies low-confidence regions for selective human review.

\subsection{Evaluation Metrics}

The evaluation protocol uses three groups of metrics corresponding to the three model outputs: segmentation metrics for the mask, topology metrics for the skeleton, and calibration metrics for the uncertainty map.

\subsubsection{Segmentation Metrics}

Let $\hat{M}_b$ and $M^{*}_b$ denote the predicted and ground truth binary masks obtained by thresholding at 0.5, respectively, where subscripts $(i,j)$ index the spatial pixel position. The four pixel-level confusion matrix entries, true positives (TP), false positives (FP), false negatives (FN), and true negatives (TN), are defined as:
\begin{equation}
    \text{TP} = \sum_{i,j} \hat{M}_{b,ij} \cdot M^{*}_{b,ij}, \quad
    \text{FP} = \sum_{i,j} \hat{M}_{b,ij} \cdot (1 - M^{*}_{b,ij}),
\end{equation}
\begin{equation}
    \text{FN} = \sum_{i,j} (1 - \hat{M}_{b,ij}) \cdot M^{*}_{b,ij}, \quad
    \text{TN} = \sum_{i,j} (1 - \hat{M}_{b,ij}) \cdot (1 - M^{*}_{b,ij}).
\end{equation}

From these, the following metrics are computed. The Dice coefficient measures the overlap between predicted and ground truth crack regions:
\begin{equation}
    \text{Dice} = \frac{2 \cdot \text{TP}}{2 \cdot \text{TP} + \text{FP} + \text{FN}}.
\end{equation}

The Intersection over Union (IoU) provides a stricter overlap measure:
\begin{equation}
    \text{IoU} = \frac{\text{TP}}{\text{TP} + \text{FP} + \text{FN}}.
\end{equation}

Precision and Recall capture the trade-off between false positives and false negatives:
\begin{equation}
    \text{Precision} = \frac{\text{TP}}{\text{TP} + \text{FP}}, \quad
    \text{Recall} = \frac{\text{TP}}{\text{TP} + \text{FN}},
\end{equation}
and the F1-score is their harmonic mean:
\begin{equation}
    \text{F1} = \frac{2 \cdot \text{Precision} \cdot \text{Recall}}{\text{Precision} + \text{Recall}}.
\end{equation}

Because crack pixels typically constitute a small fraction of the image, standard accuracy can be misleadingly high. The Matthews Correlation Coefficient (MCC) provides a balanced measure that accounts for all four confusion matrix entries:
\begin{equation}
    \text{MCC} = \frac{\text{TP} \cdot \text{TN} - \text{FP} \cdot \text{FN}}{\sqrt{(\text{TP}+\text{FP})(\text{TP}+\text{FN})(\text{TN}+\text{FP})(\text{TN}+\text{FN})}}.
\end{equation}

\subsubsection{Topology Metrics}

Standard segmentation metrics treat each pixel independently and do not reflect whether the predicted mask preserves crack connectivity and geometric structure. Two topology-aware metrics are used to evaluate the skeleton output and the structural quality of predictions.

The Skeleton Dice (SkelDice) measures overlap between the predicted skeleton $\hat{S}_b$ (thresholded at 0.5) and the ground truth skeleton $S^{*}_{\text{pruned}}$:
\begin{equation}
    \text{SkelDice} = \frac{2 \sum_{i,j} \hat{S}_{b,ij} \cdot S^{*}_{\text{pruned},ij}}{\sum_{i,j} \hat{S}_{b,ij} + \sum_{i,j} S^{*}_{\text{pruned},ij} + \epsilon},
\end{equation}
where $\epsilon$ is a small smoothing constant for numerical stability.

The centerline Dice (clDice)~\cite{shit2021cldice} evaluates topology preservation by computing the overlap between the skeletonized prediction and the skeletonized ground truth. Let $\text{Skel}(\cdot)$ denote the morphological skeletonization operator. The topology precision $T_{\text{prec}}$ and topology recall $T_{\text{rec}}$ are:
\begin{equation}
    T_{\text{prec}} = \frac{\sum_{i,j} \text{Skel}(\hat{M}_b)_{ij} \cdot M^{*}_{b,ij}}{\sum_{i,j} \text{Skel}(\hat{M}_b)_{ij} + \epsilon}, \quad
    T_{\text{rec}} = \frac{\sum_{i,j} \text{Skel}(M^{*}_b)_{ij} \cdot \hat{M}_{b,ij}}{\sum_{i,j} \text{Skel}(M^{*}_b)_{ij} + \epsilon},
\end{equation}
and the clDice is their harmonic mean:
\begin{equation}
    \text{clDice} = \frac{2 \cdot T_{\text{prec}} \cdot T_{\text{rec}}}{T_{\text{prec}} + T_{\text{rec}}}.
\end{equation}
A high clDice indicates that the prediction maintains the same connectivity and branching structure as the ground truth, even if the exact pixel boundaries differ slightly.

\subsubsection{Uncertainty Calibration Metrics}

Two metrics assess whether the predicted uncertainty $\hat{U}$ provides meaningful confidence estimates.

The Expected Calibration Error (ECE) measures the agreement between predicted confidence and actual accuracy. Predictions are partitioned into $K$ equally-spaced confidence bins $\{B_1, \ldots, B_K\}$ based on the mask probability $\hat{M}_{ij}$. The ECE is:
\begin{equation}
    \text{ECE} = \sum_{k=1}^{K} \frac{|B_k|}{n} \bigl|\text{acc}(B_k) - \text{conf}(B_k)\bigr|,
\end{equation}
where $n = H \times W$ is the total number of pixels, $|B_k|$ is the number of pixels in bin $B_k$, $\text{acc}(B_k)$ is the fraction of correctly classified pixels in that bin, and $\text{conf}(B_k)$ is the mean predicted probability. Lower ECE indicates better calibration.

The Uncertainty-Error Correlation (UncCorr) measures the Pearson correlation between per-pixel uncertainty $\hat{U}_{ij}$ and per-pixel prediction error $e_{ij} = |\hat{M}_{b,ij} - M^{*}_{b,ij}|$:
\begin{equation}
    \text{UncCorr} = \frac{\text{Cov}(\hat{U},\, e)}{\sigma_{\hat{U}} \cdot \sigma_{e}},
\end{equation}
where $\text{Cov}(\cdot, \cdot)$ denotes the covariance and $\sigma_{\hat{U}}$, $\sigma_{e}$ denote the standard deviations of the uncertainty and error maps respectively.
A high positive correlation indicates that the model assigns greater uncertainty to pixels where it makes errors, which is the desired behavior for confidence-aware decision support in engineering inspection.

\section{Experimental Setup}
 
This section describes the datasets, evaluation protocol, baseline methods, and implementation details used to evaluate CrackGeoFM.
 
\subsection{Datasets}
 
To train and evaluate a crack-domain foundation model that generalizes across infrastructure types and imaging conditions, a diverse multi-domain dataset collection is assembled from 20 publicly available crack segmentation benchmarks. The datasets span five surface materials (asphalt, concrete, masonry, stone, and mixed), four imaging platforms (smartphone, vehicle-mounted camera, UAV, and specialized industrial systems), and a range of environmental conditions including daylight, low-light, and laser illumination. Table~\ref{tab:datasets} summarizes the characteristics of all datasets.
 
\begin{table*}[t]
\centering
\caption{Summary of the 20 crack segmentation datasets used in this paper. The \textit{Role} column indicates whether each dataset is used for training (T), validation (V), or zero-shot evaluation (Z). The \textit{Crack ratio} column reports the mean percentage of crack pixels per image, highlighting the severe class imbalance common in crack segmentation. All images are resized to $448 \times 448$ during training.}
\label{tab:datasets}
\resizebox{\textwidth}{!}{%
\begin{tabular}{llccclllc}
\hline
\textbf{Dataset} & \textbf{Role} & \textbf{Train} & \textbf{Val} & \textbf{Total} & \textbf{Surface type} & \textbf{Imaging platform} & \textbf{Original resolution} & \textbf{Crack ratio (\%)} \\
\hline
\multicolumn{9}{l}{\textit{Training pool (used for model training and/or validation)}} \\
\hline
CFD~\cite{shi2016automatic}            & T+V & 100   & 18    & 118   & Asphalt pavement   & Smartphone (iPhone 5)    & 448$\times$448$^{\dagger}$  & 1.55 \\
CRACK500~\cite{zhang2016road}           & T+V & 2,858 & 505   & 3,363 & Asphalt pavement   & Smartphone               & 448$\times$448$^{\dagger}$  & 4.03 \\
DeepCrack~\cite{zou2019deepcrack}       & T+V & 443   & 78    & 521   & Asphalt, concrete  & Handheld camera          & 448$\times$448$^{\dagger}$  & 3.80 \\
forest~\cite{yang2020fphbn}             & T+V & 100   & 18    & 118   & Asphalt pavement   & Handheld camera          & 448$\times$448              & 1.54 \\
GAPS384~\cite{eisenbach2017how}         & T+V & 433   & 76    & 509   & Asphalt pavement   & Vehicle-mounted camera   & 448$\times$448$^{\dagger}$  & 1.16 \\
Sylvie~\cite{amhaz2016automatic}        & T+V & 157   & 28    & 185   & Asphalt pavement   & Road survey vehicle      & 448$\times$448$^{\dagger}$  & 8.23 \\
Volker~\cite{yang2020fphbn}             & T+V & 842   & 148   & 990   & Concrete facades   & Handheld camera          & 448$\times$448              & 4.14 \\
CrackLS315~\cite{zou2019deepcrack}      & T   & 315   & --    & 315   & Asphalt pavement   & Line-array camera (laser)& 512$\times$512              & 0.25 \\
CRKWH100~\cite{zou2019deepcrack}        & T   & 100   & --    & 100   & Asphalt pavement   & Line-array camera          & 512$\times$512              & 0.36 \\
EdmCrack600~\cite{edmcrack600}          & T   & 600   & --    & 600   & Asphalt pavement   & Vehicle-mounted (GoPro)  & 1080$\times$1920            & 0.72 \\
SUT-Crack~\cite{sutcrack2023}           & T   & 90    & --    & 90    & Asphalt pavement   & Smartphone               & 3024$\times$4032            & 1.35 \\
AEL~\cite{amhaz2016automatic}           & T   & 58    & --    & 58    & Asphalt pavement   & Road survey vehicle      & Mixed$^{\ddagger}$          & 2.42 \\
Concrete-Crack~\cite{conccrack}         & T   & 446   & --    & 446   & Concrete buildings & Close-range camera       & 3024$\times$4032            & 1.34 \\
PaveCrack1300~\cite{pavecrack1300}      & T   & 1,300 & --    & 1,300 & Asphalt pavement   & UAV (DJI Mini 4 Pro)     & 512$\times$512              & 9.70 \\
\hline
\multicolumn{9}{l}{\textit{Held-out pool (used exclusively for zero-shot and few-shot evaluation)}} \\
\hline
Eugen\_Miller~\cite{yang2020fphbn}       & Z   & --    & 55    & 55    & Tunnel lining      & Industrial camera        & 448$\times$448              & 4.91 \\
Rissbilder~\cite{yang2020fphbn}         & Z   & --    & 2,350 & 2,350 & Masonry, facades   & Handheld camera          & 448$\times$448              & 3.20 \\
Stone331~\cite{zou2019deepcrack}        & Z   & --    & 331   & 331   & Stone surfaces     & Area-array camera        & 1024$\times$1024            & 0.11 \\
cracktree200~\cite{zou2012cracktree}    & Z   & --    & 206   & 206   & Asphalt pavement   & Handheld camera          & 448$\times$448$^{\dagger}$  & 0.33 \\
LCSD~\cite{lcsd}                        & Z   & --    & 143   & 143   & Concrete structure & UAV inspection (low-light)& 400$\times$400              & 3.27 \\
UAV\_Crack~\cite{uavcrack}              & Z   & --    & 315   & 315   & Dams, levees       & UAV (high-resolution)    & 448$\times$448              & 1.38 \\
\hline
\multicolumn{9}{l}{\textbf{Total: 20 datasets, 7,842 training images, 4,271 evaluation images}} \\
\multicolumn{9}{l}{$^{\dagger}$Resized from higher original resolution during preprocessing. $^{\ddagger}$Multiple original resolutions (462$\times$311, 462$\times$991, 512$\times$768).} \\
\hline
\end{tabular}%
}
\end{table*}
 
The 20 datasets are divided into two pools based on the principle of domain diversity. The training pool comprises 14 datasets (7,842 images) that collectively cover asphalt pavement, concrete members, and concrete facades across smartphone, handheld, vehicle-mounted, and laser imaging conditions. Seven of these datasets have predefined train/val splits and are used for both training and validation; the remaining seven are flat datasets without predefined splits and are used for training only. The held-out pool comprises 6 datasets (3,400 images) that are completely excluded from training and reserved for zero-shot cross-domain and few-shot adaptation evaluation. Each held-out dataset tests a specific transfer challenge that the training pool does not cover well: Eugen\_Miller introduces tunnel imagery with unique lighting; Rissbilder provides masonry and architectural facades absent from training; Stone331 tests controlled-imaging stone surfaces; cracktree200 introduces thin-line annotation conventions with extremely sparse foreground (0.33\% crack ratio); LCSD tests low-light concrete inspection; and UAV\_Crack evaluates UAV-captured structural imagery on large infrastructure.
 
The crack ratio column in Table~\ref{tab:datasets} highlights the severe class imbalance inherent in crack segmentation. Crack pixels constitute as little as 0.11\% (Stone331) to at most 9.70\% (PaveCrack1300) of the image area, with a median around 1.5\% across datasets. This imbalance motivates the use of Dice loss and temperature-based sampling in the training protocol, as standard cross-entropy would be dominated by the overwhelming non-crack background.
 
This split ensures that the training pool provides broad material and condition coverage for learning general crack representations, while the held-out pool tests whether those representations transfer to genuinely unseen domains, materials, cameras, and annotation styles.

\subsection{Evaluation Protocol}
 
The evaluation follows a three-tier protocol designed to provide comprehensive evidence for the foundation model claim.
 
\subsubsection{Tier 1: In-domain evaluation} 
All models are evaluated on the validation splits of the seven training datasets that have predefined splits (CFD, CRACK500, DeepCrack, forest, GAPS384, Sylvie, and Volker). This tier measures standard segmentation performance on data distributions seen during training and provides the primary comparison surface against baseline methods.
 
\subsubsection{Tier 2: Zero-shot cross-domain transfer} 
All models are evaluated directly on the six held-out datasets without any fine-tuning. This is the strongest test of a foundation model because it measures whether the learned representations generalize to unseen infrastructure types, materials, and imaging conditions. All baseline models are trained on the same 14-dataset training pool, ensuring a fair comparison in which the only variable is the model architecture.
 
\subsubsection{Tier 3: Few-shot adaptation} 
For each held-out dataset, the model is fine-tuned using only $k \in \{1, 5, 10, 25\}$ labeled images sampled from the target domain, with only the CFAM adapters, FPN, and SMTD decoder heads updated while the frozen backbone and FCEM remain fixed. The remaining images serve as the test set. This tier demonstrates whether CrackGeoFM can be efficiently adapted to new inspection contexts with minimal annotation, which is the practical value proposition of a foundation model for infrastructure inspection.
 
\subsection{Baseline Methods}
CrackGeoFM is compared against 11 state-of-the-art segmentation architectures spanning CNN-based and Transformer-based designs. The CNN-based baselines include U-Net~\cite{ronneberger2015unet}, U-Net++~\cite{zhou2018unetpp}, DeepLabV3~\cite{chen2017deeplabv3}, DeepLabV3+~\cite{chen2018deeplabv3plus}, PSPNet~\cite{zhao2017pspnet}, FPN~\cite{lin2017fpn}, LinkNet~\cite{chaurasia2017linknet}, PAN~\cite{li2018pan}, MAnet~\cite{fan2020manet}, SegFormer~\cite{xie2021segformer}, and UPerNet~\cite{xiao2018upernet}, all using a ResNet-50 encoder with ImageNet pretraining. SegFormer is implemented through the \texttt{segmentation\_models\_pytorch} library with a ResNet-50 encoder rather than its original MiT backbone, ensuring a controlled comparison where the encoder is held constant across all baselines.  All baselines predict mask-only outputs (single channel), while CrackGeoFM additionally produces skeleton and uncertainty maps.
 
All baseline models are trained using the identical protocol: the same 14 training datasets with temperature-based sampling ($T = 2.0$), the same augmentation pipeline, AdamW optimizer ($\eta = 1 \times 10^{-4}$), cosine annealing schedule with 5-epoch warmup, early stopping with patience of 20 epochs, and images resized to $448 \times 448$. The only variable across experiments is the model architecture.

\section{Results and Discussion}
This section evaluates CrackGeoFM across eight experimental 
dimensions: in-domain segmentation, zero-shot transfer, 
few-shot adaptation, topology preservation, uncertainty 
calibration, ablation analysis, statistical significance, 
and computational complexity.

\subsection{In-Domain Segmentation Performance}

Table~\ref{tab:val_results} reports the per-dataset segmentation performance on the seven validation datasets. CrackGeoFM achieves the highest mean Dice (0.6843) and MCC (0.6901), surpassing U-Net++ (0.6771 Dice, 0.6784 MCC) by +0.7\% and +1.2\% respectively, and ranks first on four of the seven datasets (CFD, CRACK500, GAPS384, and Volker). No baseline dominates uniformly: U-Net++ leads on forest, U-Net on DeepCrack, and MAnet on Sylvie. CrackGeoFM avoids this inconsistency because the frozen DINOv2-Large backbone provides domain-agnostic features while the CFAM adapters learn a shared crack-domain transformation rather than dataset-specific patterns.

CrackGeoFM achieves the highest mean recall (0.7578) at slightly lower precision (0.6973) than U-Net++ (0.7154). This reflects the uncertainty-aware mask loss, which preserves ambiguous predictions with calibrated confidence rather than suppressing them through a hard decision boundary. In infrastructure inspection, missed cracks carry greater cost than flagged ambiguities, making higher recall with calibrated uncertainty preferable to higher precision with overconfident predictions.

The modest in-domain gains are expected when all models share the same training data and protocol. The value of a foundation model lies in retaining performance when the deployment domain diverges from training conditions.

\begin{figure}[H]
    \centering
    \includegraphics[width=0.8\textwidth]{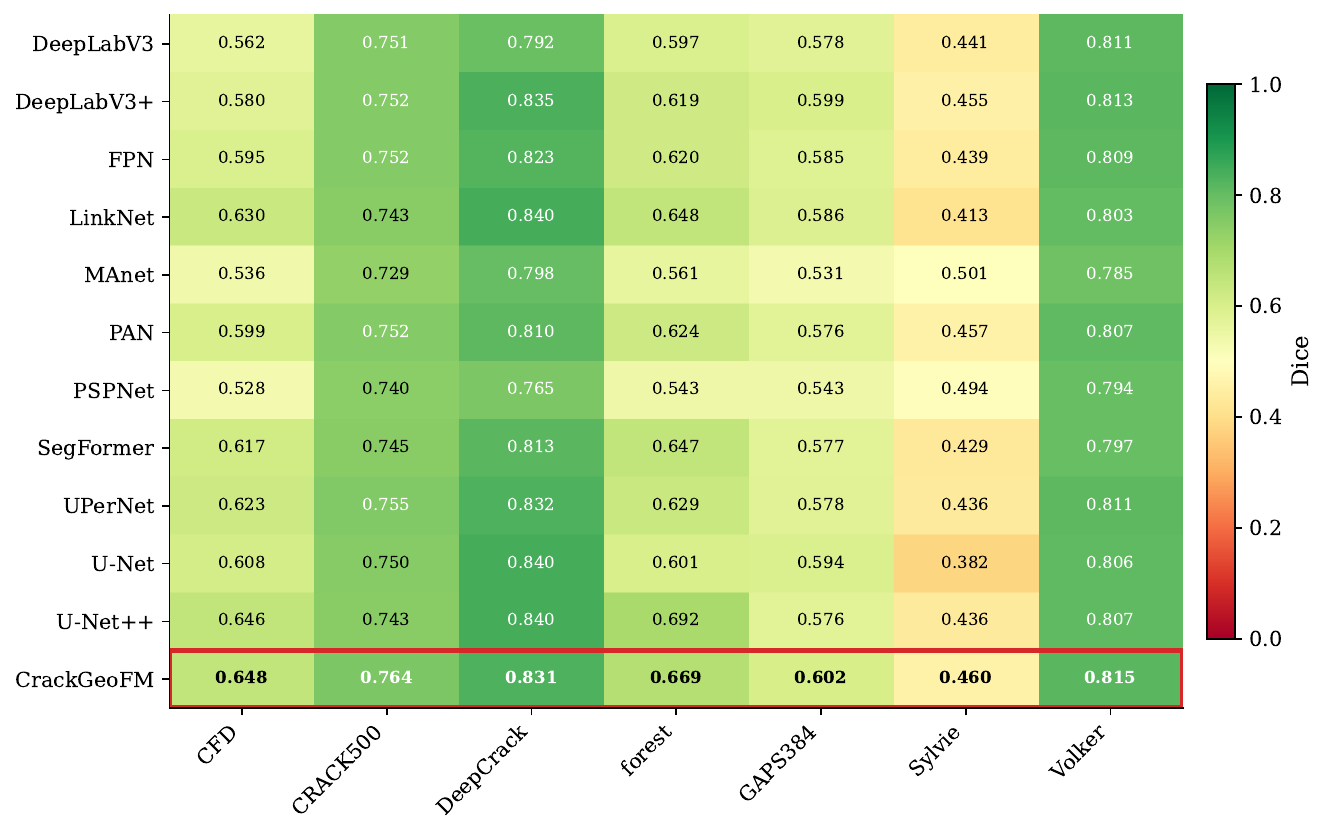}
    \caption{Per-dataset Dice heatmap on validation datasets}
    \label{fig:heatmap_val}
\end{figure}

The per-dataset heatmap (Figure~\ref{fig:heatmap_val}) 
confirms that CrackGeoFM's advantage is distributed 
across datasets rather than driven by outlier performance 
on any single domain. 

% ============================================================
% TABLE: Validation results (in-domain)
% ============================================================
\begin{table}[H]
\centering
\caption{In-domain segmentation performance on seven validation datasets. Five metrics are reported per dataset: Dice coefficient, Intersection over Union (IoU), Precision (Prec), Recall (Rec), and Matthews Correlation Coefficient (MCC). Bold indicates best; underline indicates second best. All baselines use ResNet-50 encoders. CrackGeoFM uses a frozen DINOv2-Large backbone with FCEM, CFAM, and SMTD modules.}
\label{tab:val_results}
 
% Sub-table 1: CFD, CRACK500, DeepCrack
\resizebox{\textwidth}{!}{%
\begin{tabular}{lccccccccccccccc}
\toprule
 & \multicolumn{5}{c}{\textbf{CFD}} & \multicolumn{5}{c}{\textbf{CRACK500}} & \multicolumn{5}{c}{\textbf{DeepCrack}} \\
\cmidrule(lr){2-6} \cmidrule(lr){7-11} \cmidrule(lr){12-16}
\textbf{Model} & \textbf{Dice$\uparrow$} & \textbf{IoU$\uparrow$} & \textbf{Prec$\uparrow$} & \textbf{Rec$\uparrow$} & \textbf{MCC$\uparrow$} & \textbf{Dice$\uparrow$} & \textbf{IoU$\uparrow$} & \textbf{Prec$\uparrow$} & \textbf{Rec$\uparrow$} & \textbf{MCC$\uparrow$} & \textbf{Dice$\uparrow$} & \textbf{IoU$\uparrow$} & \textbf{Prec$\uparrow$} & \textbf{Rec$\uparrow$} & \textbf{MCC$\uparrow$} \\
\midrule
U-Net & 0.6078 & 0.4546 & 0.4689 & 0.9336 & 0.6434 & 0.7499 & 0.6218 & 0.7388 & \underline{0.7942} & 0.7447 & \textbf{0.8405} & \textbf{0.7358} & 0.8082 & 0.8900 & \textbf{0.8399} \\
U-Net++ & \underline{0.6462} & \textbf{0.4945} & \textbf{0.5263} & 0.8995 & \underline{0.6709} & 0.7427 & 0.6179 & 0.7479 & 0.7705 & 0.7377 & 0.8401 & \underline{0.7346} & 0.8008 & \underline{0.8973} & \underline{0.8396} \\
DeepLabV3 & 0.5618 & 0.3958 & 0.4087 & 0.9211 & 0.6033 & 0.7514 & 0.6227 & 0.7415 & 0.7917 & 0.7457 & 0.7922 & 0.6773 & 0.7576 & 0.8492 & 0.7918 \\
DeepLabV3+ & 0.5797 & 0.4200 & 0.4280 & \textbf{0.9539} & 0.6234 & 0.7516 & 0.6231 & 0.7515 & 0.7831 & 0.7460 & 0.8352 & 0.7282 & 0.7979 & 0.8864 & 0.8338 \\
PSPNet & 0.5283 & 0.3635 & 0.3740 & 0.9244 & 0.5772 & 0.7395 & 0.6105 & 0.7352 & 0.7858 & 0.7361 & 0.7649 & 0.6397 & 0.7170 & 0.8467 & 0.7659 \\
FPN & 0.5946 & 0.4339 & 0.4440 & 0.9428 & 0.6338 & 0.7515 & 0.6243 & \underline{0.7583} & 0.7755 & 0.7461 & 0.8234 & 0.7130 & 0.7957 & 0.8660 & 0.8220 \\
LinkNet & 0.6295 & 0.4707 & 0.4899 & 0.9185 & 0.6581 & 0.7432 & 0.6164 & 0.7517 & 0.7674 & 0.7379 & \underline{0.8401} & 0.7343 & \underline{0.8189} & 0.8724 & 0.8381 \\
PAN & 0.5993 & 0.4373 & 0.4477 & \underline{0.9468} & 0.6381 & 0.7522 & 0.6232 & 0.7556 & 0.7812 & 0.7470 & 0.8102 & 0.7025 & 0.8009 & 0.8507 & 0.8120 \\
MAnet & 0.5356 & 0.3728 & 0.3906 & 0.8892 & 0.5769 & 0.7285 & 0.5982 & 0.7284 & 0.7738 & 0.7256 & 0.7979 & 0.6820 & 0.7507 & \textbf{0.9028} & 0.8048 \\
UPerNet & 0.6228 & 0.4626 & 0.4792 & 0.9317 & 0.6548 & \underline{0.7550} & \underline{0.6283} & 0.7577 & 0.7791 & \underline{0.7486} & 0.8316 & 0.7258 & \textbf{0.8195} & 0.8622 & 0.8314 \\
SegFormer & 0.6169 & 0.4535 & 0.4696 & 0.9285 & 0.6497 & 0.7454 & 0.6186 & 0.7525 & 0.7778 & 0.7410 & 0.8126 & 0.6978 & 0.7602 & 0.8885 & 0.8126 \\
\midrule
\textbf{CrackGeoFM (Ours)} & \textbf{0.6477} & \underline{0.4884} & \underline{0.5113} & 0.9103 & \textbf{0.6721} & \textbf{0.7640} & \textbf{0.6365} & \textbf{0.7585} & \textbf{0.8009} & \textbf{0.7590} & 0.8314 & 0.7213 & 0.7842 & 0.8954 & 0.8304 \\
\bottomrule
\end{tabular}
}% end resizebox
 
\vspace{1em}
 
% Sub-table 2: forest, GAPS384, Sylvie
\resizebox{\textwidth}{!}{%
\begin{tabular}{lccccccccccccccc}
\toprule
 & \multicolumn{5}{c}{\textbf{forest}} & \multicolumn{5}{c}{\textbf{GAPS384}} & \multicolumn{5}{c}{\textbf{Sylvie}} \\
\cmidrule(lr){2-6} \cmidrule(lr){7-11} \cmidrule(lr){12-16}
\textbf{Model} & \textbf{Dice$\uparrow$} & \textbf{IoU$\uparrow$} & \textbf{Prec$\uparrow$} & \textbf{Rec$\uparrow$} & \textbf{MCC$\uparrow$} & \textbf{Dice$\uparrow$} & \textbf{IoU$\uparrow$} & \textbf{Prec$\uparrow$} & \textbf{Rec$\uparrow$} & \textbf{MCC$\uparrow$} & \textbf{Dice$\uparrow$} & \textbf{IoU$\uparrow$} & \textbf{Prec$\uparrow$} & \textbf{Rec$\uparrow$} & \textbf{MCC$\uparrow$} \\
\midrule
U-Net & 0.6006 & 0.4514 & 0.4734 & 0.9226 & 0.6369 & 0.5939 & 0.4400 & \underline{0.6406} & 0.6061 & 0.6030 & 0.3819 & 0.2641 & 0.8606 & 0.2491 & 0.4187 \\
U-Net++ & \textbf{0.6921} & \textbf{0.5439} & \textbf{0.5905} & 0.8784 & \textbf{0.7068} & 0.5757 & 0.4260 & \textbf{0.6510} & 0.5701 & 0.5873 & 0.4359 & 0.3223 & \underline{0.8636} & 0.2315 & 0.4050 \\
DeepLabV3 & 0.5974 & 0.4341 & 0.4607 & 0.9049 & 0.6314 & 0.5782 & 0.4202 & 0.5845 & 0.6417 & 0.5901 & 0.4413 & 0.3305 & 0.8385 & 0.2554 & 0.4183 \\
DeepLabV3+ & 0.6191 & 0.4609 & 0.4748 & \textbf{0.9408} & 0.6537 & \underline{0.5993} & \underline{0.4404} & 0.6162 & 0.6436 & \underline{0.6094} & 0.4553 & 0.3338 & 0.8471 & 0.2868 & 0.4471 \\
PSPNet & 0.5433 & 0.3799 & 0.3946 & 0.9169 & 0.5885 & 0.5431 & 0.3907 & 0.5604 & 0.5996 & 0.5558 & \underline{0.4944} & \underline{0.3607} & 0.8359 & \textbf{0.3248} & \underline{0.4761} \\
FPN & 0.6204 & 0.4632 & 0.4800 & \underline{0.9288} & 0.6531 & 0.5850 & 0.4293 & 0.5874 & 0.6517 & 0.5968 & 0.4390 & 0.3190 & 0.8355 & 0.2885 & 0.4350 \\
LinkNet & 0.6483 & 0.4933 & 0.5258 & 0.9011 & 0.6726 & 0.5863 & 0.4308 & 0.5781 & \textbf{0.6625} & 0.5978 & 0.4127 & 0.2940 & 0.8577 & 0.2347 & 0.4130 \\
PAN & 0.6235 & 0.4640 & 0.4831 & 0.9234 & 0.6544 & 0.5757 & 0.4219 & 0.6108 & 0.6144 & 0.5884 & 0.4574 & 0.3365 & 0.8280 & 0.3034 & 0.4477 \\
MAnet & 0.5609 & 0.3945 & 0.4074 & 0.9243 & 0.6034 & 0.5311 & 0.3847 & 0.6202 & 0.5459 & 0.5496 & \textbf{0.5014} & \textbf{0.3668} & \textbf{0.8760} & 0.3009 & \textbf{0.4845} \\
UPerNet & 0.6295 & 0.4765 & 0.5031 & 0.9153 & 0.6596 & 0.5782 & 0.4266 & 0.6241 & 0.6410 & 0.5983 & 0.4360 & 0.3226 & 0.8506 & 0.2779 & 0.4329 \\
SegFormer & 0.6474 & 0.4864 & 0.5142 & 0.9022 & 0.6710 & 0.5767 & 0.4224 & 0.6130 & 0.6113 & 0.5887 & 0.4288 & 0.3052 & 0.8612 & 0.2498 & 0.4271 \\
\midrule
\textbf{CrackGeoFM (Ours)} & \underline{0.6689} & \underline{0.5118} & \underline{0.5350} & 0.9190 & \underline{0.6915} & \textbf{0.6021} & \textbf{0.4443} & 0.6200 & \underline{0.6535} & \textbf{0.6141} & 0.4603 & 0.3488 & 0.8501 & \underline{0.3132} & 0.4553 \\
\bottomrule
\end{tabular}
}% end resizebox
 
\vspace{1em}
 
% Sub-table 3: Volker + Mean
\resizebox{\textwidth}{!}{%
\begin{tabular}{lccccc}
\toprule
 & \multicolumn{5}{c}{\textbf{Volker}} \\
\cmidrule(lr){2-6}
\textbf{Model} & \textbf{Dice$\uparrow$} & \textbf{IoU$\uparrow$} & \textbf{Prec$\uparrow$} & \textbf{Rec$\uparrow$} & \textbf{MCC$\uparrow$} \\
\midrule
U-Net & 0.8061 & 0.6771 & 0.8162 & 0.8043 & 0.7999 \\
U-Net++ & 0.8073 & 0.6789 & \underline{0.8278} & 0.7967 & 0.8017 \\
DeepLabV3 & 0.8106 & 0.6830 & 0.7987 & \underline{0.8298} & 0.8041 \\
DeepLabV3+ & \underline{0.8129} & \underline{0.6861} & \textbf{0.8281} & 0.8026 & \underline{0.8062} \\
PSPNet & 0.7936 & 0.6608 & 0.7968 & 0.7995 & 0.7869 \\
FPN & 0.8092 & 0.6810 & 0.8091 & 0.8152 & 0.8024 \\
LinkNet & 0.8032 & 0.6737 & 0.7945 & 0.8231 & 0.7974 \\
PAN & 0.8072 & 0.6788 & 0.8246 & 0.7962 & 0.8006 \\
MAnet & 0.7849 & 0.6513 & 0.7752 & 0.8116 & 0.7799 \\
UPerNet & 0.8111 & 0.6838 & 0.8253 & 0.8024 & 0.8044 \\
SegFormer & 0.7967 & 0.6645 & 0.7748 & \textbf{0.8303} & 0.7906 \\
\midrule
\textbf{CrackGeoFM (Ours)} & \textbf{0.8154} & \textbf{0.6897} & 0.8219 & 0.8122 & \textbf{0.8083} \\
\bottomrule
\end{tabular}
}% end resizebox
 
\end{table}

\begin{figure}[H]
    \centering
    \includegraphics[width=1.0\textwidth]{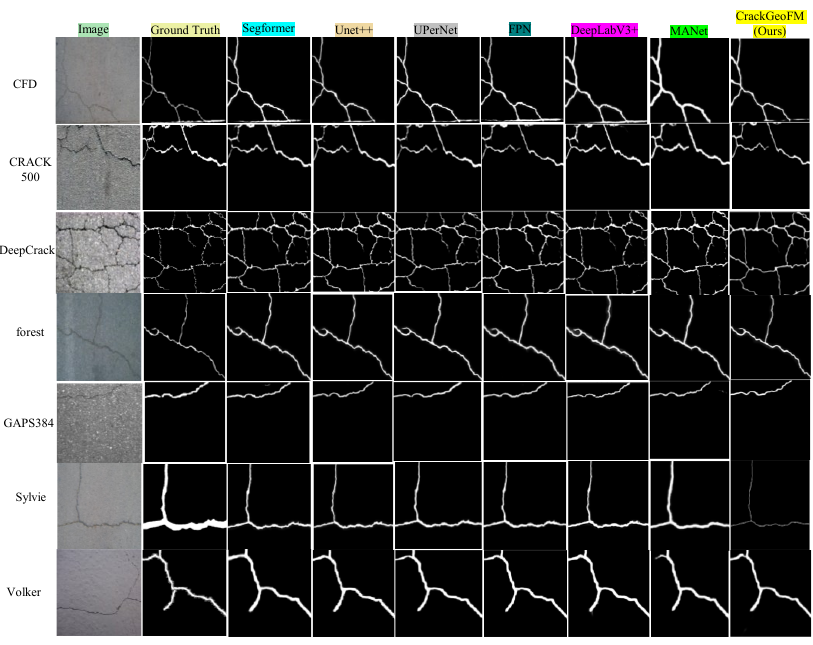}
    \caption{Qualitative comparison of segmentation predictions 
    on the seven validation datasets. Each row corresponds to a 
    representative example from the labeled dataset.}
    \label{fig:val_predictions}
\end{figure}

Figure~\ref{fig:val_predictions} provides visual confirmation: 
CrackGeoFM produces more spatially coherent predictions, 
capturing thin branches that baselines fragment (CRACK500, 
forest) and preserving network connectivity that baselines 
break into isolated segments (Sylvie, Volker). The visual 
differences become most pronounced where crack geometry is 
complex, consistent with the model's advantage stemming from 
structural coherence rather than raw pixel accuracy.

\subsection{Zero-Shot Cross-Domain Transfer}
\label{sec:zeroshot}
Table~\ref{tab:zero_results} presents the segmentation performance on six held-out datasets that were entirely excluded from training, introducing unseen infrastructure types (tunnels, masonry, dams), materials (stone, facades), and imaging conditions (low-light UAV, high-altitude inspection). CrackGeoFM achieves the highest mean Dice (0.4406), outperforming FPN (0.3957) by +4.5\% absolute, more than six times the in-domain margin. It ranks first on four of the six datasets with the highest mean IoU (0.3070), precision (0.5376), and MCC (0.4806).

The most striking result is on LCSD, the low-light UAV dataset, where CrackGeoFM achieves 0.5419 Dice while the best baseline (UPerNet) reaches only 0.1639, a margin of +37.8\%. Low illumination suppresses the intensity-based contrast that CNN baselines rely on. The FCEM addresses this by extracting high-frequency edge information through wavelet decomposition that persists even when global illumination drops, maintaining crack sensitivity where spatial intensity features fail. CrackGeoFM also outperforms all baselines on cracktree200 (+7.9\%), where extremely sparse foreground (0.33\% crack ratio) and thin-line annotations challenge all models, and on UAV\_Crack (+1.8\%), where high-altitude imagery introduces scale shifts absent from training.

The Eugen\_Miller tunnel dataset exposes a clear limitation. CrackGeoFM achieves only 0.3241 Dice versus SegFormer's 0.6776, producing very high precision (0.9286) but very low recall (0.1994). The model correctly classifies the few crack pixels it predicts but misses the majority, defaulting to high-confidence regions because the tunnel lighting and texture differ too substantially from any training domain. This conservative failure mode is preferable to overconfident false positives, and Section~\ref{sec:fewshot} demonstrates that even a few labeled tunnel images recover the recall deficit through rapid adaptation.

The generalization gap, defined as the drop in mean Dice from validation to zero-shot, quantifies transfer robustness. CrackGeoFM exhibits the smallest gap (0.244), closely followed by MAnet (0.246), then FPN (0.265), U-Net (0.286), SegFormer (0.288), U-Net++ (0.302), and DeepLabV3+ (0.311). The frozen backbone preserves broad visual priors from large-scale pretraining, the CFAM adapters learn transferable crack transformations rather than dataset-specific patterns, and the FCEM provides frequency-based features inherently more robust to appearance variation.

\begin{figure}[H]
    \centering
    \includegraphics[width=1.0\textwidth]{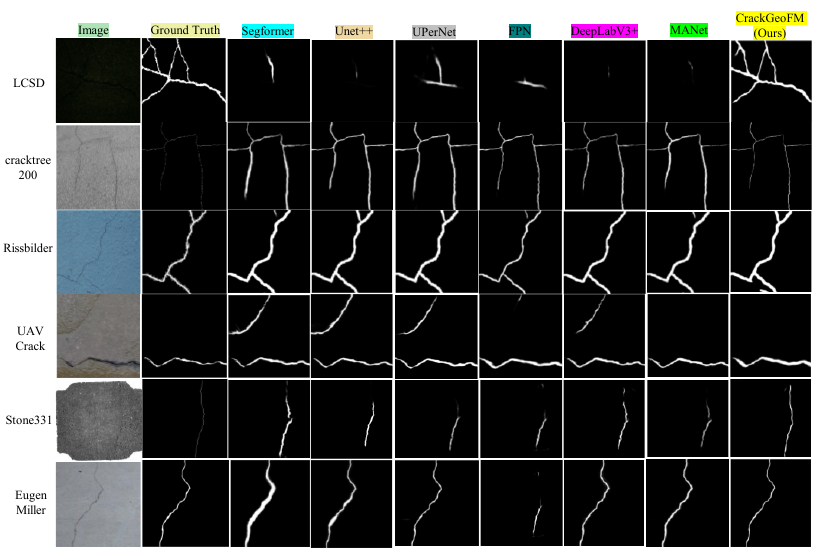}
    \caption{Qualitative segmentation comparison on six zero-shot 
    held-out datasets never seen during training.}
    \label{fig:zeroshot_predictions}
\end{figure}

Figure~\ref{fig:zeroshot_predictions} visualizes these 
cross-domain predictions. On LCSD, baselines produce 
near-empty masks because low illumination suppresses 
intensity-based contrast, while CrackGeoFM recovers the 
crack structure through FCEM's wavelet-derived edge cues 
that persist regardless of brightness. On Rissbilder and 
UAV\_Crack, the frozen backbone's domain-agnostic features 
combined with CFAM's crack specialization yield cleaner 
boundaries with fewer false positives.

\begin{figure}[H]
    \centering
    \includegraphics[width=1.0\textwidth]{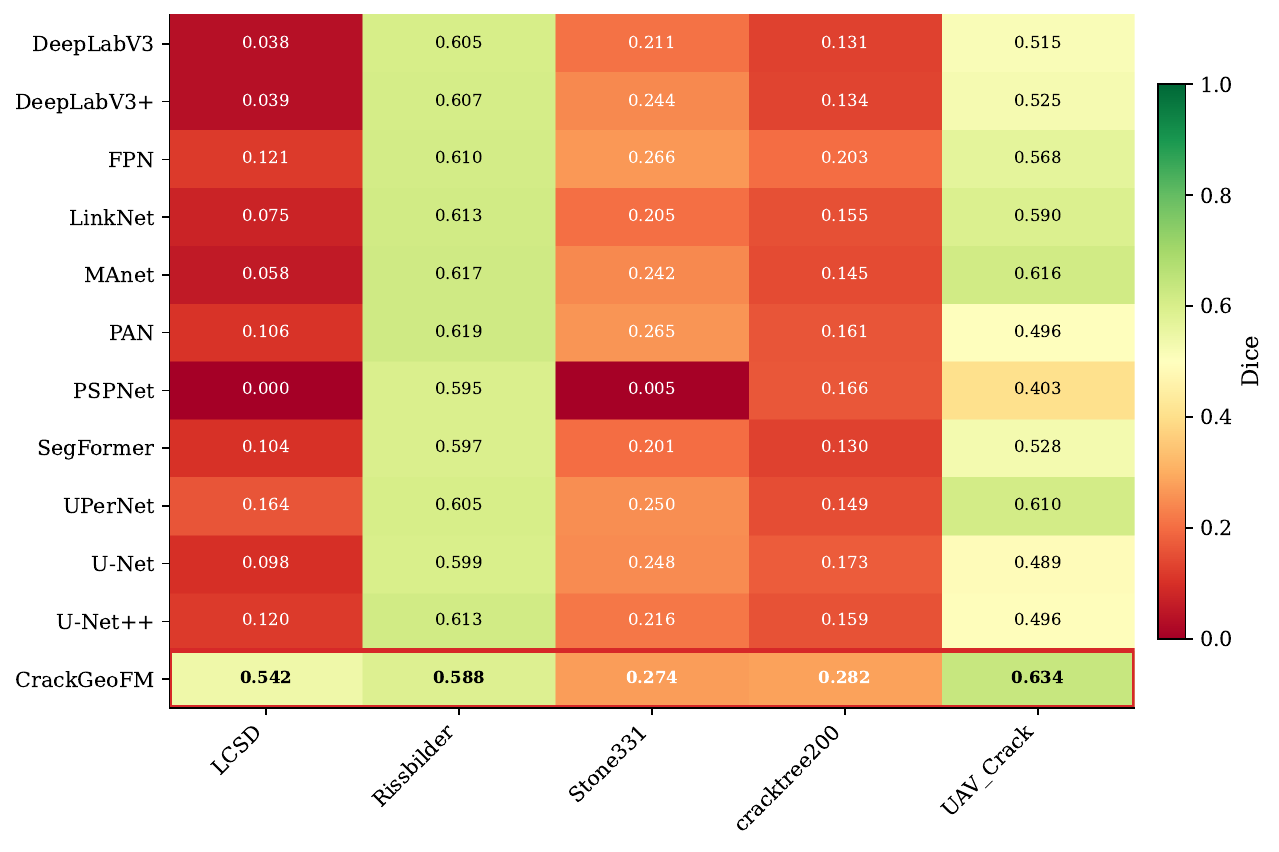}
    \caption{Per-dataset Dice heatmap on some zero-shot datasets}
    \label{fig:heatmap_zeroshot}
\end{figure}

\begin{figure}[H]
    \centering
    \includegraphics[width=0.62\textwidth]{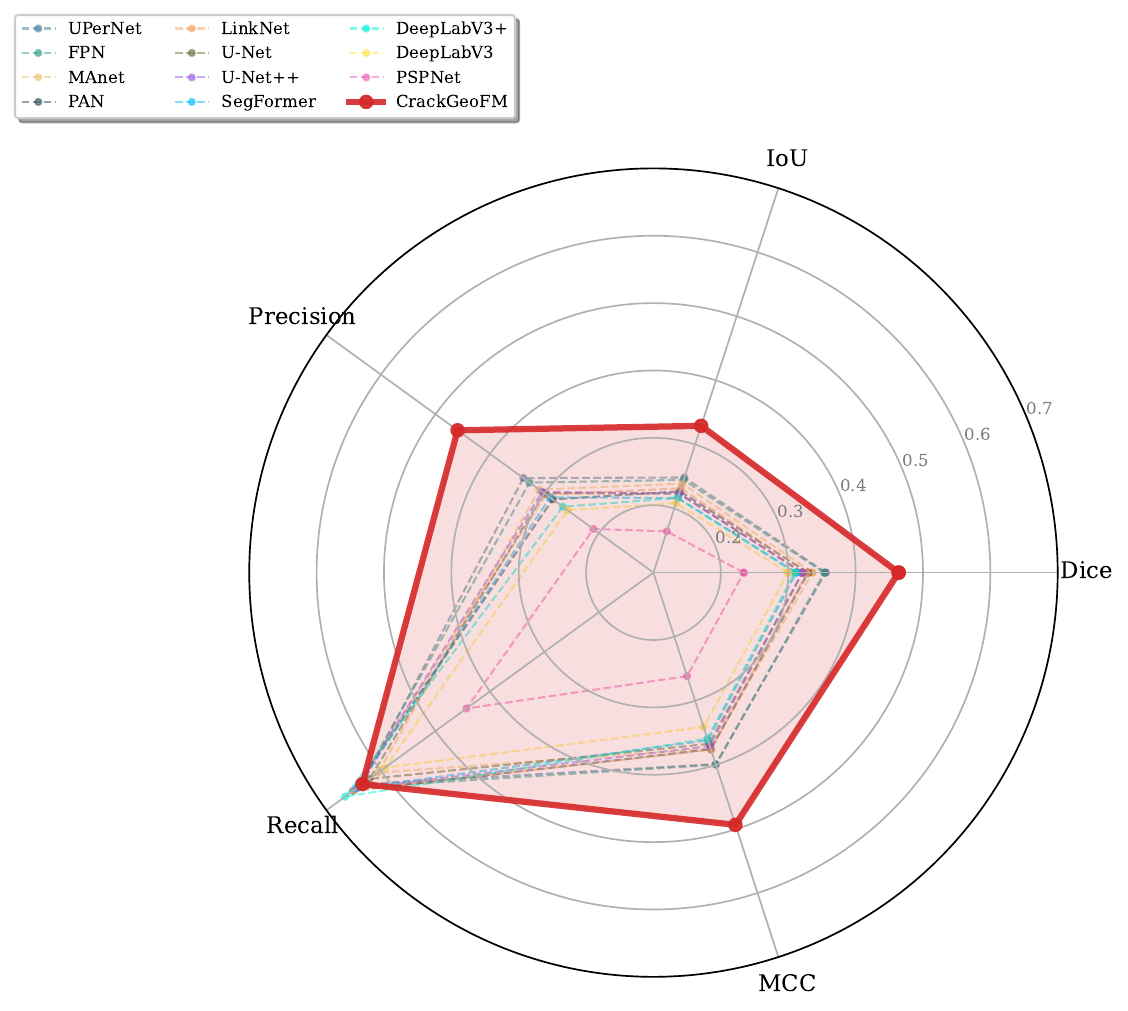}
    \caption{Multi-metric radar comparison on zero-shot datasets (mean across all six held-out datasets)}
    \label{fig:radar_zeroshot}
\end{figure}

The zero-shot heatmap (Figure~\ref{fig:heatmap_zeroshot}) illustrates the severity of domain shift, with CrackGeoFM retaining the highest values across most datasets, most notably on LCSD where all baselines approach zero while CrackGeoFM achieves 0.542. Figure~\ref{fig:radar_zeroshot} summarizes mean zero-shot performance across all five metrics: CrackGeoFM's polygon encloses a visibly larger area than all baselines, with the most pronounced advantage on Recall and Dice.

% ============================================================
% TABLE: Zero-shot results (cross-domain)
% ============================================================
\begin{table}[H]
\centering
\caption{Zero-shot cross-domain segmentation performance on six held-out datasets never seen during training. Bold indicates best; underline indicates second best.}
\label{tab:zero_results}
 
% Sub-table 1: Eugen_Miller, LCSD, Rissbilder
\resizebox{\textwidth}{!}{%
\begin{tabular}{lccccccccccccccc}
\toprule
 & \multicolumn{5}{c}{\textbf{Eugen\_Miller}} & \multicolumn{5}{c}{\textbf{LCSD}} & \multicolumn{5}{c}{\textbf{Rissbilder}} \\
\cmidrule(lr){2-6} \cmidrule(lr){7-11} \cmidrule(lr){12-16}
\textbf{Model} & \textbf{Dice$\uparrow$} & \textbf{IoU$\uparrow$} & \textbf{Prec$\uparrow$} & \textbf{Rec$\uparrow$} & \textbf{MCC$\uparrow$} & \textbf{Dice$\uparrow$} & \textbf{IoU$\uparrow$} & \textbf{Prec$\uparrow$} & \textbf{Rec$\uparrow$} & \textbf{MCC$\uparrow$} & \textbf{Dice$\uparrow$} & \textbf{IoU$\uparrow$} & \textbf{Prec$\uparrow$} & \textbf{Rec$\uparrow$} & \textbf{MCC$\uparrow$} \\
\midrule
U-Net & 0.6029 & 0.4475 & 0.6362 & 0.6497 & 0.6022 & 0.0984 & 0.0649 & 0.2039 & 0.0861 & 0.1126 & 0.5985 & 0.4405 & 0.5762 & 0.6767 & 0.5949 \\
U-Net++ & 0.6459 & 0.4839 & 0.6382 & \underline{0.7138} & 0.6419 & 0.1201 & 0.0784 & 0.2496 & 0.1103 & 0.1362 & 0.6133 & 0.4535 & 0.5762 & 0.7040 & 0.6104 \\
DeepLabV3 & 0.5458 & 0.3875 & 0.7605 & 0.5139 & 0.5747 & 0.0382 & 0.0243 & 0.0605 & 0.0356 & 0.0414 & 0.6046 & 0.4454 & 0.5643 & 0.7012 & 0.6026 \\
DeepLabV3+ & 0.5765 & 0.4129 & 0.7556 & 0.5410 & 0.5971 & 0.0390 & 0.0255 & 0.1321 & 0.0410 & 0.0465 & 0.6067 & 0.4448 & 0.5221 & \underline{0.7529} & 0.6066 \\
PSPNet & 0.6048 & 0.4476 & 0.7666 & 0.5283 & 0.6111 & 0.0002 & 0.0002 & 0.0000 & 0.0000 & 0.0000 & 0.5953 & 0.4355 & 0.5618 & 0.6796 & 0.5930 \\
FPN & 0.6067 & 0.4412 & 0.7625 & 0.5688 & 0.6214 & 0.1206 & 0.0769 & 0.2954 & 0.1034 & 0.1434 & 0.6101 & 0.4496 & 0.5842 & 0.6808 & 0.6058 \\
LinkNet & 0.4566 & 0.3004 & \underline{0.8789} & 0.3140 & 0.5084 & 0.0755 & 0.0485 & 0.1470 & 0.0779 & 0.0862 & 0.6135 & 0.4533 & 0.5574 & 0.7229 & 0.6098 \\
PAN & 0.5120 & 0.3620 & 0.8582 & 0.4062 & 0.5550 & 0.1061 & 0.0667 & 0.1864 & 0.1099 & 0.1167 & \textbf{0.6192} & \textbf{0.4571} & 0.5539 & 0.7267 & \textbf{0.6146} \\
MAnet & \underline{0.6523} & \underline{0.4867} & 0.8061 & 0.5646 & \underline{0.6557} & 0.0582 & 0.0377 & 0.1566 & 0.0495 & 0.0698 & \underline{0.6173} & \underline{0.4554} & 0.5502 & 0.7372 & \underline{0.6135} \\
UPerNet & 0.5148 & 0.3583 & 0.7983 & 0.4582 & 0.5541 & \underline{0.1639} & \underline{0.1023} & \underline{0.3030} & \underline{0.1400} & \underline{0.1783} & 0.6054 & 0.4442 & \underline{0.5986} & 0.6672 & 0.6059 \\
SegFormer & \textbf{0.6776} & \textbf{0.5186} & 0.6546 & \textbf{0.7178} & \textbf{0.6655} & 0.1037 & 0.0676 & 0.2387 & 0.0938 & 0.1214 & 0.5968 & 0.4358 & 0.5103 & \textbf{0.7553} & 0.5976 \\
\midrule
\textbf{CrackGeoFM (Ours)} & 0.3241 & 0.1966 & \textbf{0.9286} & 0.1994 & 0.4156 & \textbf{0.5419} & \textbf{0.3970} & \textbf{0.7167} & \textbf{0.4857} & \textbf{0.5565} & 0.5878 & 0.4265 & \textbf{0.6021} & 0.6613 & 0.5963 \\
\bottomrule
\end{tabular}
}% end resizebox
 
\vspace{1em}
 
% Sub-table 2: Stone331, cracktree200, UAV_Crack_Dataset
\resizebox{\textwidth}{!}{%
\begin{tabular}{lccccccccccccccc}
\toprule
 & \multicolumn{5}{c}{\textbf{Stone331}} & \multicolumn{5}{c}{\textbf{cracktree200}} & \multicolumn{5}{c}{\textbf{UAV\_Crack}} \\
\cmidrule(lr){2-6} \cmidrule(lr){7-11} \cmidrule(lr){12-16}
\textbf{Model} & \textbf{Dice$\uparrow$} & \textbf{IoU$\uparrow$} & \textbf{Prec$\uparrow$} & \textbf{Rec$\uparrow$} & \textbf{MCC$\uparrow$} & \textbf{Dice$\uparrow$} & \textbf{IoU$\uparrow$} & \textbf{Prec$\uparrow$} & \textbf{Rec$\uparrow$} & \textbf{MCC$\uparrow$} & \textbf{Dice$\uparrow$} & \textbf{IoU$\uparrow$} & \textbf{Prec$\uparrow$} & \textbf{Rec$\uparrow$} & \textbf{MCC$\uparrow$} \\
\midrule
U-Net & 0.2479 & 0.1434 & 0.1622 & 0.7258 & 0.3278 & 0.1727 & 0.0957 & 0.0975 & 0.8972 & 0.2836 & 0.4888 & 0.3714 & 0.4637 & 0.7194 & 0.5171 \\
U-Net++ & 0.2161 & 0.1237 & 0.1475 & 0.6756 & 0.2943 & 0.1592 & 0.0869 & 0.0874 & 0.9423 & 0.2792 & 0.4956 & 0.3757 & 0.4591 & 0.7923 & 0.5339 \\
DeepLabV3 & 0.2109 & 0.1193 & 0.1444 & 0.5605 & 0.2692 & 0.1312 & 0.0705 & 0.0778 & 0.8741 & 0.2415 & 0.5148 & 0.3820 & 0.4410 & 0.7932 & 0.5496 \\
DeepLabV3+ & 0.2438 & 0.1400 & 0.1530 & \underline{0.7522} & 0.3291 & 0.1342 & 0.0727 & 0.0736 & \underline{0.9448} & 0.2504 & 0.5245 & 0.3994 & 0.4511 & \underline{0.8385} & 0.5620 \\
PSPNet & 0.0046 & 0.0028 & 0.0535 & 0.0020 & 0.0086 & 0.1662 & 0.0909 & 0.0985 & 0.7327 & 0.2545 & 0.4034 & 0.2926 & 0.3373 & 0.8013 & 0.4519 \\
FPN & \underline{0.2658} & \underline{0.1551} & 0.1731 & 0.6903 & 0.3359 & \underline{0.2027} & \underline{0.1132} & \underline{0.1161} & 0.8709 & \underline{0.3092} & 0.5682 & 0.4287 & 0.4678 & \textbf{0.8646} & 0.6048 \\
LinkNet & 0.2052 & 0.1156 & 0.1261 & \textbf{0.7965} & 0.3040 & 0.1553 & 0.0848 & 0.0859 & 0.9068 & 0.2699 & 0.5903 & 0.4554 & 0.5686 & 0.7412 & 0.6124 \\
PAN & 0.2654 & 0.1542 & 0.1745 & 0.6988 & \underline{0.3374} & 0.1608 & 0.0879 & 0.0900 & 0.9160 & 0.2761 & 0.4963 & 0.3684 & 0.4195 & 0.8025 & 0.5334 \\
MAnet & 0.2425 & 0.1417 & \underline{0.1790} & 0.5639 & 0.2964 & 0.1454 & 0.0791 & 0.0804 & 0.8839 & 0.2551 & \underline{0.6160} & \underline{0.4803} & \underline{0.5811} & 0.7915 & \underline{0.6385} \\
UPerNet & 0.2504 & 0.1465 & 0.1781 & 0.5949 & 0.3075 & 0.1486 & 0.0808 & 0.0815 & \textbf{0.9481} & 0.2683 & 0.6104 & 0.4684 & 0.5322 & 0.8207 & 0.6343 \\
SegFormer & 0.2013 & 0.1135 & 0.1306 & 0.6909 & 0.2835 & 0.1298 & 0.0698 & 0.0702 & 0.9270 & 0.2460 & 0.5281 & 0.3961 & 0.4991 & 0.7517 & 0.5600 \\
\midrule
\textbf{CrackGeoFM (Ours)} & \textbf{0.2739} & \textbf{0.1611} & \textbf{0.1887} & 0.6739 & \textbf{0.3403} & \textbf{0.2821} & \textbf{0.1649} & \textbf{0.1934} & 0.5594 & \textbf{0.3213} & \textbf{0.6339} & \textbf{0.4958} & \textbf{0.5959} & 0.7898 & \textbf{0.6538} \\
\bottomrule
\end{tabular}
}% end resizebox
 
\end{table}

\subsection{Few-Shot Adaptation}
\label{sec:fewshot}

The zero-shot results reveal strong transfer on most held-out datasets but also limitations on domains that differ substantially from training, such as tunnel imagery. Few-shot adaptation tests whether minimal labeled data can bridge this gap. For each held-out dataset, $k \in \{1, 5, 10, 25\}$ images are sampled for fine-tuning while the rest serve as the test set. Only the CFAM adapters, FPN, and decoder heads are updated; the backbone and FCEM remain frozen. Each experiment is repeated with three random seeds.

Table~\ref{tab:fewshot} presents the results. Mean Dice improves from 0.4627 (zero-shot) to 0.5585 with five labeled images (+9.6\%) and 0.5890 with 25 images (+12.6\%), with most gain occurring by $k{=}5$. The most dramatic case is Eugen\_Miller, where Dice rises from 0.3673 to 0.7857 ($k{=}25$, +41.8\%), directly resolving the conservative failure mode identified in Section~\ref{sec:zeroshot}: the adapters learn domain-specific adjustments that recover recall without sacrificing precision. UAV\_Crack shows a similar trajectory (0.6338 to 0.7423, +10.9\%).

At $k{=}1$, three of six datasets show a slight drop compared to zero-shot because a single unrepresentative image can mislead the adapters. By $k{=}5$, all datasets surpass zero-shot performance, establishing five labeled images as a reliable deployment minimum. For infrastructure agencies, this means labeling a handful of crack images from a new structure type suffices to achieve performance substantially above zero-shot without retraining the full model.

\begin{table}[H]
\centering
\caption{Few-shot adaptation of CrackGeoFM on held-out datasets. $k$ denotes the number of labeled support images used for fine-tuning. Only CFAM adapters, FPN, and decoder heads are updated; the frozen backbone and FCEM remain fixed. Results report Dice coefficient (mean $\pm$ std across 3 random seeds). $k{=}0$ corresponds to zero-shot evaluation on the test partition after excluding the support pool, and therefore differs slightly from the full-dataset zero-shot values in Table~\ref{tab:zero_results}.}
\label{tab:fewshot}
\resizebox{\textwidth}{!}{%
\begin{tabular}{lccccc}
\toprule
\textbf{Dataset} & $k{=}0$ (zero-shot) & $k{=}1$ & $k{=}5$ & $k{=}10$ & $k{=}25$ \\
\midrule
Eugen\_Miller & 0.3673 & 0.6909$\pm$0.0333 & 0.7549$\pm$0.0079 & 0.7738$\pm$0.0087 & 0.7857$\pm$0.0011 \\
Rissbilder & 0.5700 & 0.5664$\pm$0.0428 & 0.6178$\pm$0.0200 & 0.6331$\pm$0.0032 & 0.6391$\pm$0.0034 \\
Stone331 & 0.2736 & 0.2908$\pm$0.0149 & 0.3272$\pm$0.0125 & 0.3328$\pm$0.0077 & 0.3510$\pm$0.0008 \\
cracktree200 & 0.2872 & 0.2529$\pm$0.0088 & 0.2857$\pm$0.0114 & 0.2983$\pm$0.0006 & 0.3099$\pm$0.0018 \\
LCSD & 0.6441 & 0.5666$\pm$0.0181 & 0.6539$\pm$0.0110 & 0.6910$\pm$0.0050 & 0.7060$\pm$0.0063 \\
UAV\_Crack & 0.6338 & 0.6923$\pm$0.0196 & 0.7116$\pm$0.0240 & 0.7319$\pm$0.0095 & 0.7423$\pm$0.0100 \\
\midrule
\textbf{Mean} & \textbf{0.4627} & \textbf{0.5100} & \textbf{0.5585} & \textbf{0.5768} & \textbf{0.5890} \\
\midrule
$\Delta$ vs.\ zero-shot & -- & +0.0473 & +0.0958 & +0.1141 & +0.1263 \\
\bottomrule
\end{tabular}%
}
\end{table}

\begin{figure}[H]
    \centering
    \includegraphics[width=0.8\textwidth]{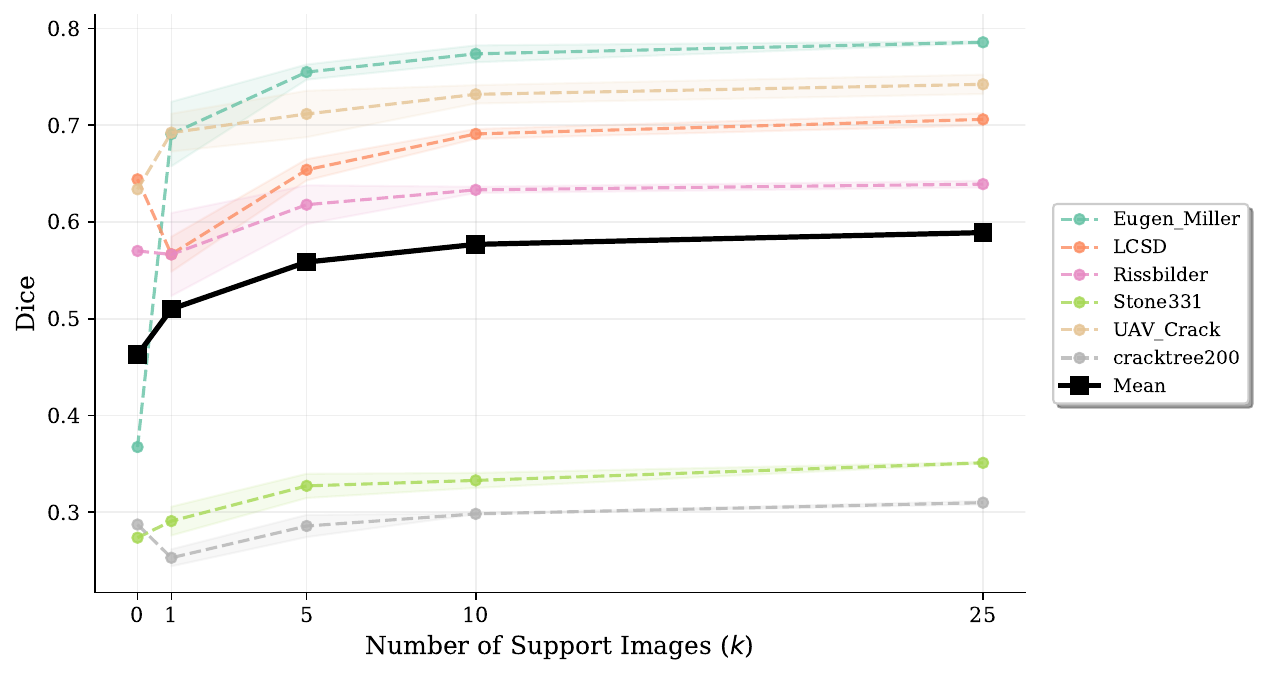}
    \caption{Few-shot adaptation curves on six held-out datasets.}
    \label{fig:fewshot_curve}
\end{figure}

Figure~\ref{fig:fewshot_curve} shows the adaptation 
trajectory for each dataset, with most curves plateauing 
by $k{=}5$. Eugen\_Miller exhibits the steepest climb 
(0.37 to 0.79), confirming that a few labeled examples 
suffice to activate the frozen backbone's latent 
crack-relevant features for unseen domains.

\subsection{Topology Preservation}
Tables~\ref{tab:topology_val} and~\ref{tab:topology_zero} compare skeleton quality across all models. Baseline skeletons are obtained by applying the same morphological pipeline (closing, Zhang-Suen thinning, pruning) to predicted masks; CrackGeoFM skeletons come directly from the skeleton head.

On validation datasets, CrackGeoFM's direct skeleton achieves 0.2896 mean SkelDice, outperforming the best baseline post-hoc (U-Net++, 0.2402, +4.9\%) and its own post-hoc (0.2423, +4.7\%). The gap between direct and post-hoc from the same model is the key finding: both start from the same mask, but the skeleton head produces a more accurate centerline because it is trained with explicit topology supervision rather than relying on morphological extraction from a mask that may contain breaks, noise, or false connections.

On zero-shot datasets, the advantage persists: direct achieves 0.2139 SkelDice versus the best baseline post-hoc (0.1499, +6.4\%) and own post-hoc (0.1691, +4.5\%). The skeleton head learns structural priors about crack connectivity that generalize across domains, whereas post-hoc skeletonization inherits every topological error in the predicted mask.

% ============================================================
% TABLE: Topology comparison - Validation (per-dataset)
% ============================================================
\begin{table*}[t]
\centering
\caption{Topology preservation on validation datasets: post-hoc skeletonization of predicted masks versus direct skeleton prediction by CrackGeoFM. SD = SkelDice; clD = clDice. All post-hoc skeletons use the same morphological pipeline. Bold indicates best; underline indicates second best.}
\label{tab:topology_val}

\resizebox{\textwidth}{!}{%
\begin{tabular}{llcccccccccccccccc}
\toprule
 &  & \multicolumn{2}{c}{\textbf{CFD}} & \multicolumn{2}{c}{\textbf{CRACK500}} & \multicolumn{2}{c}{\textbf{DeepCrack}} & \multicolumn{2}{c}{\textbf{forest}} & \multicolumn{2}{c}{\textbf{GAPS384}} & \multicolumn{2}{c}{\textbf{Sylvie}} & \multicolumn{2}{c}{\textbf{Volker}} & \multicolumn{2}{c}{\textbf{Mean}} \\
\cmidrule(lr){3-4} \cmidrule(lr){5-6} \cmidrule(lr){7-8} \cmidrule(lr){9-10} \cmidrule(lr){11-12} \cmidrule(lr){13-14} \cmidrule(lr){15-16} \cmidrule(lr){17-18}
\textbf{Model} & \textbf{Skel.} & \textbf{SD$\uparrow$} & \textbf{clD$\uparrow$} & \textbf{SD$\uparrow$} & \textbf{clD$\uparrow$} & \textbf{SD$\uparrow$} & \textbf{clD$\uparrow$} & \textbf{SD$\uparrow$} & \textbf{clD$\uparrow$} & \textbf{SD$\uparrow$} & \textbf{clD$\uparrow$} & \textbf{SD$\uparrow$} & \textbf{clD$\uparrow$} & \textbf{SD$\uparrow$} & \textbf{clD$\uparrow$} & \textbf{SD$\uparrow$} & \textbf{clD$\uparrow$} \\
\midrule
U-Net & P-hoc & 0.2767 & 0.8172 & 0.1125 & 0.8271 & \underline{0.3706} & 0.8951 & 0.2647 & 0.8157 & 0.2382 & 0.7091 & 0.1375 & 0.6692 & 0.1613 & 0.7545 & 0.2231 & 0.7840 \\
U-Net++ & P-hoc & 0.2965 & 0.8315 & 0.1154 & 0.8201 & 0.3580 & \textbf{0.8998} & 0.3027 & 0.8519 & 0.2231 & 0.6878 & \textbf{0.2188} & 0.7289 & 0.1667 & 0.7469 & 0.2402 & 0.7953 \\
DeepLabV3 & P-hoc & 0.2276 & 0.8076 & 0.1058 & 0.8304 & 0.2693 & 0.8418 & 0.2164 & 0.8173 & 0.1978 & 0.6898 & 0.1723 & 0.6810 & \underline{0.1728} & \underline{0.7753} & 0.1946 & 0.7776 \\
DeepLabV3+ & P-hoc & 0.2816 & 0.8264 & 0.1128 & 0.8304 & 0.3428 & 0.8899 & 0.2937 & 0.8406 & 0.2361 & 0.7144 & 0.1696 & 0.7214 & 0.1648 & 0.7382 & 0.2288 & 0.7945 \\
PSPNet & P-hoc & 0.2431 & 0.8145 & 0.1059 & 0.8175 & 0.2486 & 0.8205 & 0.2376 & 0.8207 & 0.1772 & 0.6408 & 0.1828 & \underline{0.7595} & 0.1562 & 0.7499 & 0.1931 & 0.7748 \\
FPN & P-hoc & 0.2595 & 0.8075 & 0.1144 & 0.8274 & 0.3144 & 0.8795 & 0.2812 & 0.8206 & 0.2226 & 0.7002 & 0.1698 & 0.7139 & 0.1635 & 0.7612 & 0.2179 & 0.7872 \\
LinkNet & P-hoc & 0.2894 & 0.8369 & 0.1142 & 0.8168 & 0.3549 & \underline{0.8952} & 0.2949 & 0.8382 & 0.2321 & 0.7065 & 0.1779 & 0.7069 & 0.1694 & 0.7550 & 0.2333 & 0.7936 \\
PAN & P-hoc & 0.2756 & 0.8331 & 0.1112 & 0.8272 & 0.3200 & 0.8635 & 0.2893 & 0.8430 & 0.2189 & 0.6904 & 0.1759 & 0.7194 & 0.1647 & 0.7344 & 0.2222 & 0.7873 \\
MAnet & P-hoc & 0.2432 & 0.8050 & 0.1049 & 0.8080 & 0.3363 & 0.8896 & 0.2628 & 0.8434 & 0.2220 & 0.6485 & \underline{0.1996} & \textbf{0.7837} & 0.1569 & 0.7711 & 0.2179 & 0.7928 \\
UPerNet & P-hoc & 0.2915 & 0.8386 & \underline{0.1171} & 0.8295 & 0.3635 & 0.8851 & 0.2771 & 0.8353 & \underline{0.2506} & 0.7019 & 0.1834 & 0.7125 & 0.1673 & 0.7431 & 0.2358 & 0.7923 \\
SegFormer & P-hoc & 0.2820 & 0.8500 & 0.1115 & 0.8223 & 0.3070 & 0.8755 & 0.2948 & 0.8566 & 0.2052 & 0.6828 & 0.1719 & 0.7196 & 0.1542 & \textbf{0.7868} & 0.2181 & 0.7991 \\
\midrule
CrackGeoFM (Ours) & P-hoc & \underline{0.3273} & \textbf{0.8566} & \textbf{0.1171} & \textbf{0.8459} & 0.3436 & 0.8937 & \underline{0.3309} & \textbf{0.8796} & 0.2315 & \textbf{0.7223} & 0.1821 & 0.7278 & 0.1640 & 0.7453 & \underline{0.2423} & \textbf{0.8102} \\
\textbf{CrackGeoFM (Ours)} & \textbf{Direct} & \textbf{0.3992} & \textbf{0.8566} & 0.0985 & \textbf{0.8459} & \textbf{0.3980} & 0.8937 & \textbf{0.4147} & \textbf{0.8796} & \textbf{0.2880} & \textbf{0.7223} & 0.1672 & 0.7278 & \textbf{0.2615} & 0.7453 & \textbf{0.2896} & \textbf{0.8102} \\
\bottomrule
\end{tabular}
}% end resizebox

\end{table*}

% ============================================================
% TABLE: Topology comparison - Zero-shot (per-dataset)
% ============================================================
\begin{table}[H]
\centering
\caption{Topology preservation on zero-shot datasets: post-hoc skeletonization versus direct skeleton prediction. Bold indicates best; underline indicates second best.}
\label{tab:topology_zero}

\resizebox{\textwidth}{!}{%
\begin{tabular}{llcccccccccccccc}
\toprule
 &  & \multicolumn{2}{c}{\textbf{Eugen\_Miller}} & \multicolumn{2}{c}{\textbf{LCSD}} & \multicolumn{2}{c}{\textbf{Rissbilder}} & \multicolumn{2}{c}{\textbf{Stone331}} & \multicolumn{2}{c}{\textbf{cracktree200}} & \multicolumn{2}{c}{\textbf{UAV\_Crack}} & \multicolumn{2}{c}{\textbf{Mean}} \\
\cmidrule(lr){3-4} \cmidrule(lr){5-6} \cmidrule(lr){7-8} \cmidrule(lr){9-10} \cmidrule(lr){11-12} \cmidrule(lr){13-14} \cmidrule(lr){15-16}
\textbf{Model} & \textbf{Skel.} & \textbf{SD$\uparrow$} & \textbf{clD$\uparrow$} & \textbf{SD$\uparrow$} & \textbf{clD$\uparrow$} & \textbf{SD$\uparrow$} & \textbf{clD$\uparrow$} & \textbf{SD$\uparrow$} & \textbf{clD$\uparrow$} & \textbf{SD$\uparrow$} & \textbf{clD$\uparrow$} & \textbf{SD$\uparrow$} & \textbf{clD$\uparrow$} & \textbf{SD$\uparrow$} & \textbf{clD$\uparrow$} \\
\midrule
U-Net & P-hoc & 0.1069 & 0.6452 & 0.0472 & 0.1258 & 0.1647 & 0.6697 & 0.2349 & \textbf{0.5085} & 0.0888 & 0.3091 & 0.2444 & 0.6286 & 0.1478 & 0.4811 \\
U-Net++ & P-hoc & 0.1088 & \underline{0.7019} & 0.0468 & 0.1512 & \underline{0.1650} & 0.6884 & \underline{0.2397} & \underline{0.5065} & 0.0904 & 0.3146 & 0.2489 & 0.6594 & 0.1499 & \textbf{0.5037} \\
DeepLabV3 & P-hoc & 0.0999 & 0.5254 & 0.0180 & 0.0489 & 0.1566 & 0.6831 & 0.1249 & 0.3095 & 0.0764 & 0.2591 & 0.2095 & 0.6671 & 0.1142 & 0.4155 \\
DeepLabV3+ & P-hoc & 0.1016 & 0.5440 & 0.0177 & 0.0505 & 0.1514 & 0.7021 & 0.2197 & 0.4787 & 0.0832 & 0.2888 & 0.2453 & 0.6757 & 0.1365 & 0.4567 \\
PSPNet & P-hoc & 0.1094 & 0.5183 & 0.0000 & 0.0000 & 0.1461 & 0.6555 & 0.0017 & 0.0032 & 0.0673 & 0.2394 & 0.1718 & 0.5548 & 0.0827 & 0.3285 \\
FPN & P-hoc & 0.1087 & 0.5650 & 0.0383 & 0.1455 & 0.1550 & 0.6463 & 0.2021 & 0.4509 & 0.0772 & 0.2892 & 0.2504 & 0.7440 & 0.1386 & 0.4735 \\
LinkNet & P-hoc & 0.0984 & 0.3510 & 0.0314 & 0.1015 & 0.1633 & 0.6897 & 0.2176 & 0.4623 & 0.0822 & 0.2895 & \underline{0.3008} & 0.7177 & 0.1489 & 0.4353 \\
PAN & P-hoc & 0.0919 & 0.4269 & 0.0357 & 0.1435 & 0.1571 & 0.6889 & 0.1977 & 0.4455 & 0.0872 & 0.3046 & 0.2337 & 0.6362 & 0.1339 & 0.4409 \\
MAnet & P-hoc & \underline{0.1170} & 0.5722 & 0.0343 & 0.0745 & 0.1604 & \textbf{0.7232} & 0.1771 & 0.3996 & 0.0908 & \underline{0.3167} & 0.2948 & 0.7409 & 0.1458 & 0.4712 \\
UPerNet & P-hoc & 0.1052 & 0.4826 & 0.0546 & 0.2076 & 0.1523 & 0.6429 & 0.1723 & 0.3944 & \underline{0.0926} & \textbf{0.3226} & 0.2445 & 0.7515 & 0.1369 & 0.4669 \\
SegFormer & P-hoc & 0.0965 & \textbf{0.7095} & 0.0495 & 0.1300 & 0.1509 & \underline{0.7217} & 0.1592 & 0.3650 & 0.0800 & 0.2998 & 0.2379 & 0.6778 & 0.1290 & 0.4840 \\
\midrule
CrackGeoFM (Ours) & P-hoc & 0.0952 & 0.2384 & \textbf{0.2016} & \textbf{0.6359} & 0.1569 & 0.6434 & 0.1788 & 0.4115 & \textbf{0.0974} & 0.3038 & 0.2848 & \textbf{0.7691} & \underline{0.1691} & \underline{0.5004} \\
\textbf{CrackGeoFM (Ours)} & \textbf{Direct} & \textbf{0.1563} & 0.2384 & \underline{0.1920} & \textbf{0.6359} & \textbf{0.2481} & 0.6434 & \textbf{0.3008} & 0.4115 & 0.0553 & 0.3038 & \textbf{0.3309} & \textbf{0.7691} & \textbf{0.2139} & \underline{0.5004} \\
\bottomrule
\end{tabular}
}% end resizebox

\end{table}

The clDice results show that skeleton supervision also improves the mask itself. CrackGeoFM achieves 0.8102 mean clDice versus SegFormer's 0.7991 on validation data, confirming that the skeleton loss backpropagates through the shared representation and encourages the mask to preserve crack connectivity. This validates the mask-derived supervision strategy: generating skeleton targets from existing annotations and training a dedicated head produces better topology than post-hoc extraction, while simultaneously improving mask quality through multi-task learning.

\begin{figure}[H]
    \centering
    \includegraphics[width=1.0\textwidth]{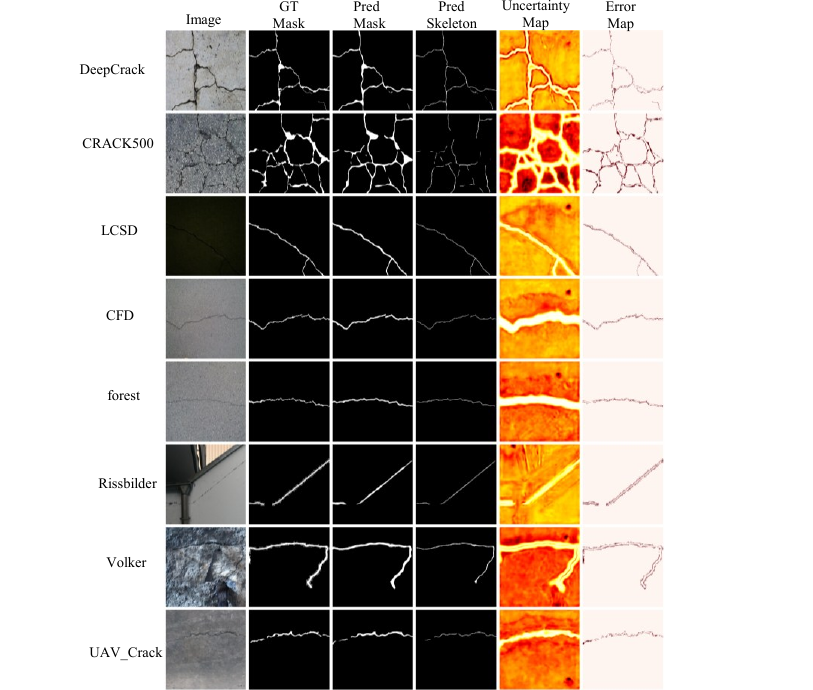}
    \caption{CrackGeoFM's three simultaneous outputs (Pred Mask, Pred Skeleton, Uncertainty Map) across 
    eight datasets.}
    \label{fig:multi_output}
\end{figure}

Figure~\ref{fig:multi_output} visualizes all three outputs 
across eight representative datasets. The predicted mask and 
skeleton capture crack geometry consistently across both 
in-domain (DeepCrack, CRACK500, CFD) and zero-shot (LCSD, 
Rissbilder, UAV\_Crack) datasets. The uncertainty map uses a 
hot colormap where dark red indicates high confidence and 
bright yellow indicates high uncertainty; the error map shows 
misclassified pixels in dark red against a light background.
Across all examples, the bright regions in the uncertainty 
map align spatially with the dark regions in the error map, 
confirming that the uncertainty head learns to identify 
crack boundaries and ambiguous textures where prediction 
errors concentrate. This spatial correspondence is 
particularly visible on DeepCrack (network junctions) and 
Volker (thin linear cracks), where the model assigns peak 
uncertainty precisely at the transition zones between crack 
and background.

\subsection{Uncertainty Calibration}
Table~\ref{tab:uncertainty_calibration} reports ECE and UncCorr for CrackGeoFM across all datasets. Mean ECE is 0.0187 on validation and 0.0169 on zero-shot data, indicating close agreement between predicted confidence and actual accuracy. The slight improvement on zero-shot data suggests the uncertainty head learns a generalizable relationship between visual ambiguity and prediction reliability rather than memorizing training-domain calibration patterns.

UncCorr confirms a moderate positive correlation between predicted uncertainty and actual error (0.3432 val, 0.3078 zero-shot). The correlation transfers to unseen domains, meaning the uncertainty map provides useful confidence signals even on infrastructure types the model has never encountered. Per-dataset variation is informative: among validation datasets, DeepCrack achieves the lowest ECE (0.0069) due to high-contrast cracks, while Sylvie has the highest (0.0551) due to complex textures and network crack patterns. The uncertainty head correctly identifies datasets with higher inherent ambiguity rather than producing uniform confidence regardless of difficulty.

From a deployment perspective, calibrated uncertainty enables selective human review: inspectors focus on high-uncertainty regions rather than verifying all predictions, reducing review effort while maintaining coverage of error-prone areas.

\begin{table}[H]
\centering
\caption{Uncertainty calibration of CrackGeoFM (DINOv2-Large) across all evaluation datasets. ECE (Expected Calibration Error) measures agreement between predicted confidence and actual accuracy (lower is better). UncCorr (Uncertainty-Error Correlation) measures the Pearson correlation between predicted uncertainty and actual prediction error (higher indicates the model correctly identifies where it is likely to make mistakes).}
\label{tab:uncertainty_calibration}
\begin{tabular}{lcc}
\toprule
\textbf{Dataset} & \textbf{ECE}$\downarrow$ & \textbf{UncCorr}$\uparrow$ \\
\midrule
\multicolumn{3}{l}{\textit{Validation (in-domain)}} \\
\midrule
CFD & 0.0172 & 0.2784 \\
CRACK500 & 0.0184 & 0.3580 \\
DeepCrack & 0.0069 & 0.3064 \\
GAPS384 & 0.0076 & 0.3319 \\
Sylvie & 0.0551 & 0.4785 \\
Volker & 0.0099 & 0.3711 \\
forest & 0.0161 & 0.2779 \\
\midrule
\textbf{Val Mean} & \textbf{0.0187} & \textbf{0.3432} \\
\midrule
\multicolumn{3}{l}{\textit{Zero-shot (cross-domain)}} \\
\midrule
Eugen\_Miller & 0.0372 & 0.2582 \\
LCSD & 0.0196 & 0.3590 \\
Rissbilder & 0.0189 & 0.3925 \\
Stone331 & 0.0033 & 0.1967 \\
UAV\_Crack & 0.0149 & 0.3393 \\
cracktree200 & 0.0074 & 0.3011 \\
\midrule
\textbf{ZS Mean} & \textbf{0.0169} & \textbf{0.3078} \\
\bottomrule
\end{tabular}
\end{table}

\begin{figure}[H]
    \centering
    \includegraphics[width=1.0\textwidth]{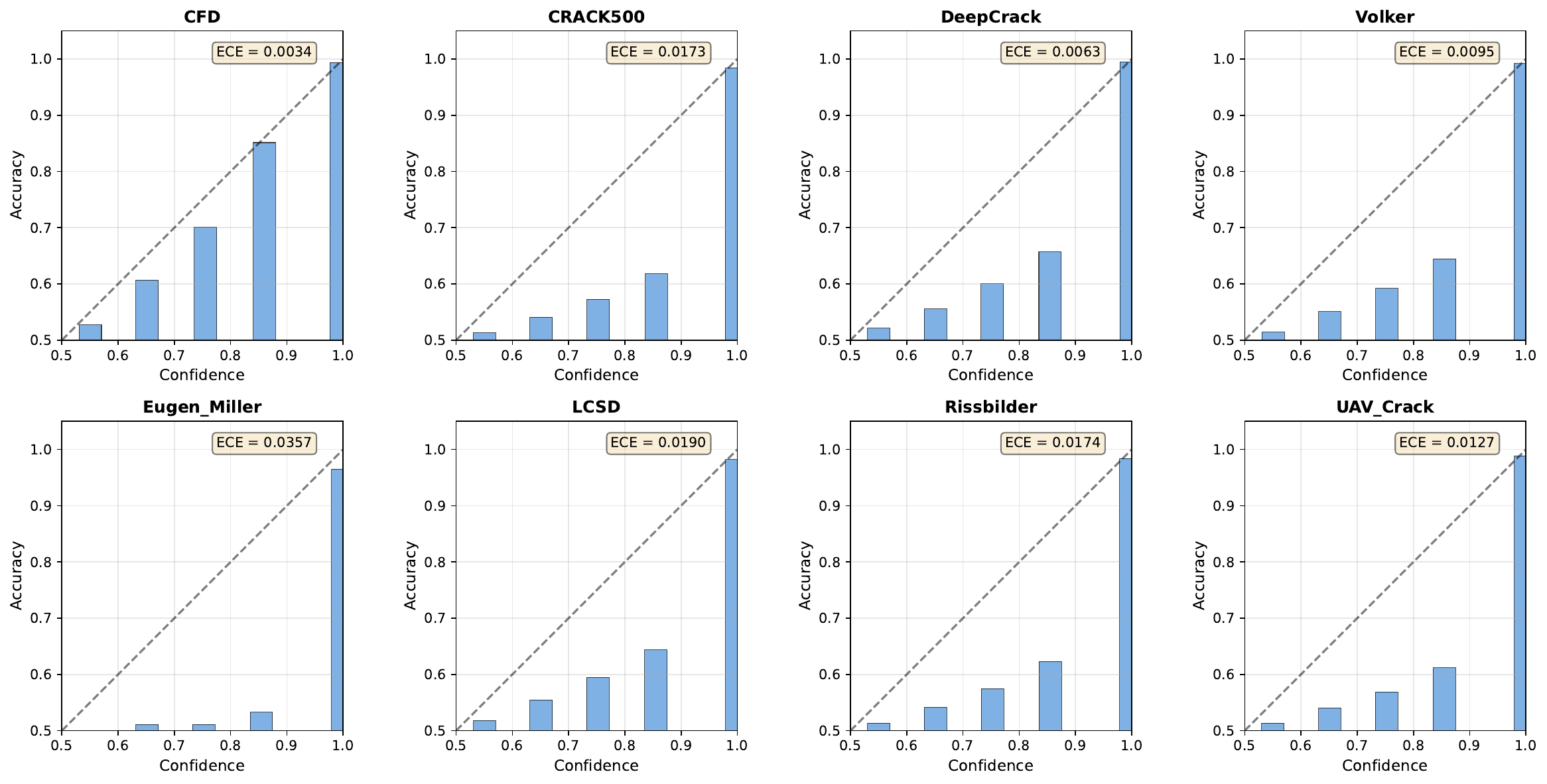}  % change file name and width as needed
    \caption{Reliability diagrams for four validation (top row) and four zero-shot (bottom row) datasets. Bars show binned accuracy versus confidence for predictions above 0.5 probability.}
    \label{fig:reliability}
\end{figure}

Figure~\ref{fig:reliability} presents reliability diagrams for four 
validation and four zero-shot datasets. The binned accuracy bars 
closely follow the perfect calibration diagonal across all datasets. At high confidence levels (0.9-1.0), predicted 
accuracy matches actual accuracy on both familiar and unseen domains, 
meaning confident predictions can be trusted for automated reporting 
while uncertain regions are routed to human review. 

% % ============================================================
% % TABLE: Topology and uncertainty per dataset
% % ============================================================
% \begin{table}[t]
% \centering
% \caption{Topology preservation and uncertainty calibration for CrackGeoFM (DINOv2-Large) across all evaluation datasets.}
% \label{tab:topology_uncertainty}
% \begin{tabular}{lcccc}
% \toprule
% \textbf{Dataset} & \textbf{SkelDice$\uparrow$} & \textbf{clDice$\uparrow$} & \textbf{ECE$\downarrow$} & \textbf{UncCorr$\uparrow$} \\
% \midrule
% \multicolumn{5}{l}{\textit{Validation (in-domain)}} \\
% \midrule
% CFD & 0.4111 & 0.9970 & 0.0172 & 0.2784 \\
% CRACK500 & 0.0854 & 0.6231 & 0.0184 & 0.3580 \\
% DeepCrack & 0.4053 & 0.9920 & 0.0069 & 0.3064 \\
% GAPS384 & 0.2941 & 0.9820 & 0.0076 & 0.3319 \\
% Sylvie & 0.2411 & 0.8922 & 0.0551 & 0.4785 \\
% Volker & 0.2852 & 0.9129 & 0.0099 & 0.3711 \\
% forest & 0.4277 & 0.9976 & 0.0161 & 0.2779 \\
% \textbf{Val Mean} & \textbf{0.3071} & \textbf{0.9138} & \textbf{0.0187} & \textbf{0.3432} \\
% \midrule
% \multicolumn{5}{l}{\textit{Zero-shot (cross-domain)}} \\
% \midrule
% Eugen\_Miller & 0.1575 & 0.8904 & 0.0372 & 0.2582 \\
% LCSD & 0.1983 & 0.8757 & 0.0196 & 0.3590 \\
% Rissbilder & 0.2667 & 0.8740 & 0.0189 & 0.3925 \\
% Stone331 & 0.3223 & 0.9408 & 0.0033 & 0.1967 \\
% UAV\_Crack\_Dataset & 0.3447 & 0.9911 & 0.0149 & 0.3393 \\
% cracktree200 & 0.0522 & 0.9137 & 0.0074 & 0.3011 \\
% \textbf{ZS Mean} & \textbf{0.2236} & \textbf{0.9143} & \textbf{0.0169} & \textbf{0.3078} \\
% \bottomrule
% \end{tabular}
% \end{table}

\subsection{Ablation Study}
\label{sec:ablation}

\subsubsection{Ablation experiment setup}
\label{sec:ablation_setup}
CrackGeoFM is evaluated through an incremental ablation study to measure the contribution of each module and loss component. Starting from a frozen DINOv2-Large backbone with an FPN and mask head, each configuration adds one component until the full model is reached.

Table~\ref{tab:ablation_configs} summarizes six settings. Config A is the baseline with BCE and Dice loss. Config B adds CFAM adapters to test lightweight domain adaptation. Config C adds FCEM to assess the value of frequency-domain crack cues. Config D adds skeleton prediction with BCE and clDice supervision. Config E adds uncertainty prediction with an uncertainty-aware mask loss. Config F adds topology consistency loss, completing the full CrackGeoFM model. All configurations are trained on the same 14-dataset training pool with identical hyperparameters and evaluated on the seven validation datasets and six zero-shot held-out datasets using the same protocol.

\begin{table}[t]
\centering
\caption{Ablation configurations. Each row adds one component to the previous configuration. CFAM = Crack-Domain Feature Adaptation Module; FCEM = Frequency-Guided Crack Enhancement Module; Skel = skeleton head with BCE + clDice loss; Unc = uncertainty head with uncertainty-aware mask loss; Consist = topology consistency loss.}
\label{tab:ablation_configs}
\resizebox{\columnwidth}{!}{%
\begin{tabular}{clccccc}
\toprule
\textbf{Config} & \textbf{Description} & \textbf{CFAM} & \textbf{FCEM} & \textbf{Skel} & \textbf{Unc} & \textbf{Consist} \\
\midrule
A & Frozen backbone + FPN + Mask   & \xmark & \xmark & \xmark & \xmark & \xmark \\
B & A + CFAM adapters              & \cmark & \xmark & \xmark & \xmark & \xmark \\
C & B + FCEM frequency injection   & \cmark & \cmark & \xmark & \xmark & \xmark \\
D & C + skeleton head              & \cmark & \cmark & \cmark & \xmark & \xmark \\
E & D + uncertainty head           & \cmark & \cmark & \cmark & \cmark & \xmark \\
\midrule
F & E + consistency loss (Full)    & \cmark & \cmark & \cmark & \cmark & \cmark \\
\bottomrule
\end{tabular}%
}
\end{table}

\subsubsection{Analysis}
Table~\ref{tab:ablation} reports the mean Dice, IoU, and MCC for each configuration on both validation and zero-shot datasets. The results reveal a non-monotonic trajectory: performance improves with CFAM, dips when skeleton supervision and frequency features are added in isolation, then recovers and surpasses all intermediate configurations when the full loss formulation is applied.
 
Adding the CFAM adapters (A$\to$B) provides the strongest single-module contribution (+2.3\% val Dice, +1.1\% zero-shot). Frozen backbone features, while broadly informative, lack specialization for crack characteristics such as extreme foreground sparsity and subtle contrast against textured backgrounds. The adapters address this with only 525K additional parameters, confirming that lightweight bottleneck adaptation is sufficient to specialize general visual features for the crack domain.
 
Introducing the FCEM frequency side-channel (B$\to$C) yields a slight decrease ($-$0.7\% val, $-$0.8\% zero-shot). In isolation, the wavelet-derived high-frequency features do not improve mask prediction because the adapter lacks structural supervision to contextualize frequency cues as crack evidence rather than texture noise. The frequency information becomes beneficial only within the full multi-task framework, where skeleton supervision provides the geometric context needed to interpret high-frequency edge responses.
 
Adding skeleton supervision (C$\to$D) causes the largest single-step drop ($-$4.5\% val, $-$1.8\% zero-shot). This is a well-documented multi-task learning effect: introducing a competing objective without an alignment mechanism creates gradient interference in the shared representation, forcing it to compromise between mask coverage and skeleton accuracy.
 
The uncertainty head (D$\to$E) partially recovers performance (+2.3\% val, +1.2\% zero-shot) by reducing the mask loss penalty on ambiguous pixels where the mask and skeleton objectives disagree, alleviating the gradient conflict introduced in Config D.
 
The topology consistency loss (E$\to$F) provides the largest improvement (+4.2\% val, +4.4\% zero-shot) and is the critical component that makes the multi-task architecture effective. By enforcing bidirectional agreement between the predicted mask and skeleton, the consistency loss aligns the two competing objectives and allows the shared representation to benefit from both supervision signals. The zero-shot gain (+4.4\%) exceeding the val gain (+4.2\%) indicates that the aligned multi-task representation generalizes better than either single-task or unaligned alternatives.
 
The total gain from baseline to full model is +3.7\% val Dice and +4.0\% zero-shot Dice, with every held-out dataset improving. The non-monotonic trajectory reveals that the proposed modules are complementary rather than independently additive: FCEM, skeleton supervision, and consistency are individually insufficient but collectively essential, with the consistency loss serving as the binding mechanism that enables multi-task crack representation learning.

\begin{table}[H]
\centering
\caption{Ablation study of CrackGeoFM components. Each row incrementally adds one module to the previous configuration. Checkmarks indicate active components. Dice, IoU, and MCC are reported as means across all datasets in each evaluation tier.}
\label{tab:ablation}
\resizebox{\textwidth}{!}{%
\begin{tabular}{clccccccccccc}
\toprule
& & \multicolumn{5}{c}{\textbf{Components}} & \multicolumn{3}{c}{\textbf{Validation}} & \multicolumn{3}{c}{\textbf{Zero-shot}} \\
\cmidrule(lr){3-7} \cmidrule(lr){8-10} \cmidrule(lr){11-13}
\textbf{Config} & \textbf{Description} & \textbf{CFAM} & \textbf{FCEM} & \textbf{Skel} & \textbf{Unc} & \textbf{Consist} & \textbf{Dice}$\uparrow$ & \textbf{IoU}$\uparrow$ & \textbf{MCC}$\uparrow$ & \textbf{Dice}$\uparrow$ & \textbf{IoU}$\uparrow$ & \textbf{MCC}$\uparrow$ \\
\midrule
A & Baseline & \xmark & \xmark & \xmark & \xmark & \xmark & 0.6477 & 0.5121 & 0.6575 & 0.4006 & 0.2759 & 0.4443 \\
B & A + CFAM & \cmark & \xmark & \xmark & \xmark & \xmark & 0.6707 & 0.5366 & 0.6747 & 0.4112 & 0.2855 & 0.4509 \\
C & B + FCEM & \cmark & \cmark & \xmark & \xmark & \xmark & 0.6641 & 0.5338 & 0.6652 & 0.4037 & 0.2813 & 0.4429 \\
D & C + Skeleton & \cmark & \cmark & \cmark & \xmark & \xmark & 0.6193 & 0.4882 & 0.6371 & 0.3853 & 0.2657 & 0.4269 \\
E & D + Uncertainty & \cmark & \cmark & \cmark & \cmark & \xmark & 0.6421 & 0.5031 & 0.6539 & 0.3969 & 0.2729 & 0.4473 \\
\midrule
F & E + Consistency (Full) & \cmark & \cmark & \cmark & \cmark & \cmark & \textbf{0.6843} & \textbf{0.5487} & \textbf{0.6901} & \textbf{0.4406} & \textbf{0.3070} & \textbf{0.4806} \\
\bottomrule
\end{tabular}%
}
\end{table}

\begin{figure}[H]
    \centering
    \includegraphics[width=1.0\textwidth]{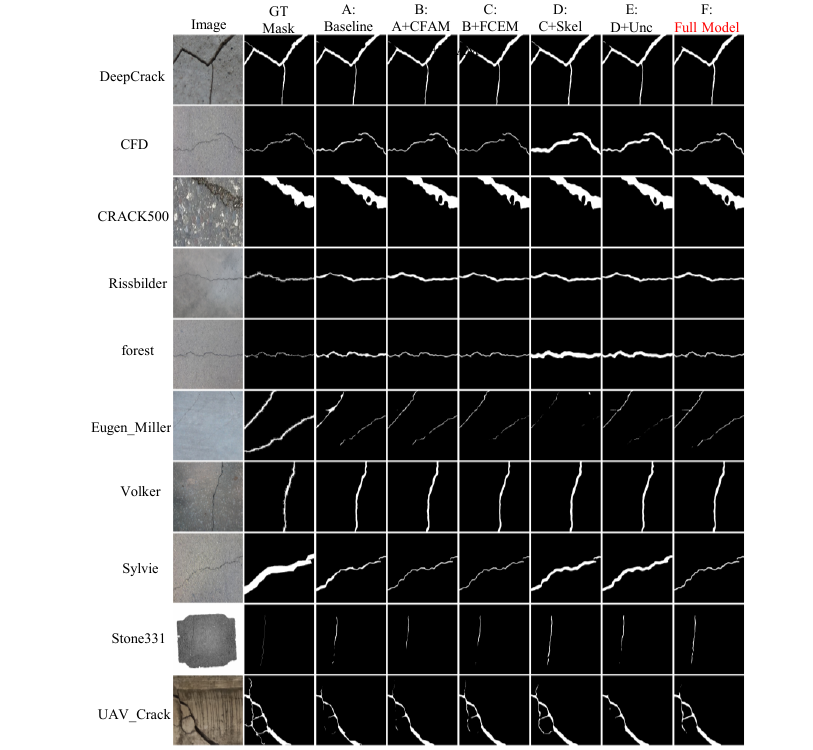}
    \caption{Predictions from each ablation configuration (A through F) 
    across in-domain and zero-shot datasets}
    \label{fig:ablation_predictions}
\end{figure}

Figure~\ref{fig:ablation_predictions} visualizes the 
incremental effect of each component. Config A (frozen 
backbone only) captures coarse crack regions but misses 
fine branches and produces fragmented predictions. Adding 
CFAM (B) visibly improves crack completeness, particularly 
on thin structures (CRACK500, forest). The intermediate 
configs (C, D, E) show the non-monotonic behavior reported 
in Table~\ref{tab:ablation}: predictions become noisier 
when skeleton supervision competes with the mask objective 
without alignment. The full model (F) resolves this, 
producing the cleanest and most connected predictions 
across all datasets, confirming that the consistency loss 
enables the shared representation to benefit from all 
supervision signals simultaneously.

\subsubsection{Backbone Ablation}
Table~\ref{tab:backbone_ablation} compares five backbones within CrackGeoFM, keeping all other components fixed. On validation data, all five perform within a 1.2\% Dice range (0.6754-0.6870), confirming that the proposed modules effectively specialize diverse backbones for crack segmentation regardless of architecture.

Zero-shot evaluation reveals a clearer separation. The four ViT backbones cluster between 0.42-0.44 Dice, while ResNet-101 drops to 0.3291. Self-supervised ViT representations transfer more robustly to unseen domains than supervised CNN features, and the proposed modules cannot fully compensate for a backbone with weaker transferable representations. Among the ViT variants, DINOv2-B/14 slightly outperforms DINOv2-L/14 on zero-shot Dice (0.4445 vs.\ 0.4406), suggesting that the smaller model regularizes better under domain shift. 

Across all backbones, uncertainty calibration remains reliable: ECE stays below 0.03 and UncCorr exceeds 0.28 for all variants, confirming that the SMTD decoder produces consistent uncertainty predictions independent of backbone choice. DINOv2-L/14 is selected as the primary model based on its balance of in-domain performance, competitive transfer, and the highest validation MCC (0.6901).

% ============================================================
% TABLE: Backbone ablation (stacked vertically)
% ============================================================
\begin{table}[H]
\centering
\caption{CrackGeoFM performance with different backbone architectures. Mean values across validation and zero-shot datasets. Bold indicates best per metric.}
\label{tab:backbone_ablation}

% --- Validation ---
\resizebox{\textwidth}{!}{%
\begin{tabular}{lccccccc}
\toprule
& \multicolumn{7}{c}{\textbf{Validation (Mean)}} \\
\cmidrule(lr){2-8}
\textbf{Backbone} & \textbf{Dice}$\uparrow$ & \textbf{IoU}$\uparrow$ & \textbf{Prec}$\uparrow$ & \textbf{Rec}$\uparrow$ & \textbf{MCC}$\uparrow$ & \textbf{ECE}$\downarrow$ & \textbf{UncCorr}$\uparrow$ \\
\midrule
DINOv2-L/14 & 0.6843 & 0.5487 & 0.6973 & \textbf{0.7578} & \textbf{0.6901} & 0.0187 & 0.3432 \\
DINOv2-B/14 & 0.6839 & 0.5499 & 0.6943 & 0.7519 & 0.6846 & 0.0186 & \textbf{0.3586} \\
DINOv3-L/16 & \textbf{0.6870} & \textbf{0.5514} & \textbf{0.7066} & 0.7386 & 0.6861 & 0.0180 & 0.3308 \\
DINOv3-B/16 & 0.6764 & 0.5383 & 0.6921 & 0.7424 & 0.6810 & \textbf{0.0176} & 0.3553 \\
ResNet-101 & 0.6754 & 0.5376 & 0.7035 & 0.7167 & 0.6743 & 0.0177 & 0.3276 \\
\bottomrule
\end{tabular}%
}

\vspace{1em}

% --- Zero-shot ---
\resizebox{\textwidth}{!}{%
\begin{tabular}{lccccccc}
\toprule
& \multicolumn{7}{c}{\textbf{Zero-shot (Mean)}} \\
\cmidrule(lr){2-8}
\textbf{Backbone} & \textbf{Dice}$\uparrow$ & \textbf{IoU}$\uparrow$ & \textbf{Prec}$\uparrow$ & \textbf{Rec}$\uparrow$ & \textbf{MCC}$\uparrow$ & \textbf{ECE}$\downarrow$ & \textbf{UncCorr}$\uparrow$ \\
\midrule
DINOv2-L/14 & 0.4406 & 0.3070 & \textbf{0.5376} & 0.5616 & 0.4806 & 0.0169 & 0.3078 \\
DINOv2-B/14 & \textbf{0.4445} & \textbf{0.3134} & 0.5365 & 0.5741 & 0.4851 & \textbf{0.0164} & \textbf{0.3697} \\
DINOv3-L/16 & 0.4430 & 0.3115 & 0.5085 & \textbf{0.6311} & \textbf{0.4882} & 0.0183 & 0.3368 \\
DINOv3-B/16 & 0.4179 & 0.2923 & 0.4981 & 0.6123 & 0.4663 & 0.0201 & 0.3608 \\
ResNet-101 & 0.3291 & 0.2234 & 0.3290 & 0.5802 & 0.3723 & 0.0276 & 0.2879 \\
\bottomrule
\end{tabular}%
}

\end{table}

\begin{figure}[H]
    \centering
    \includegraphics[width=1.0\textwidth]{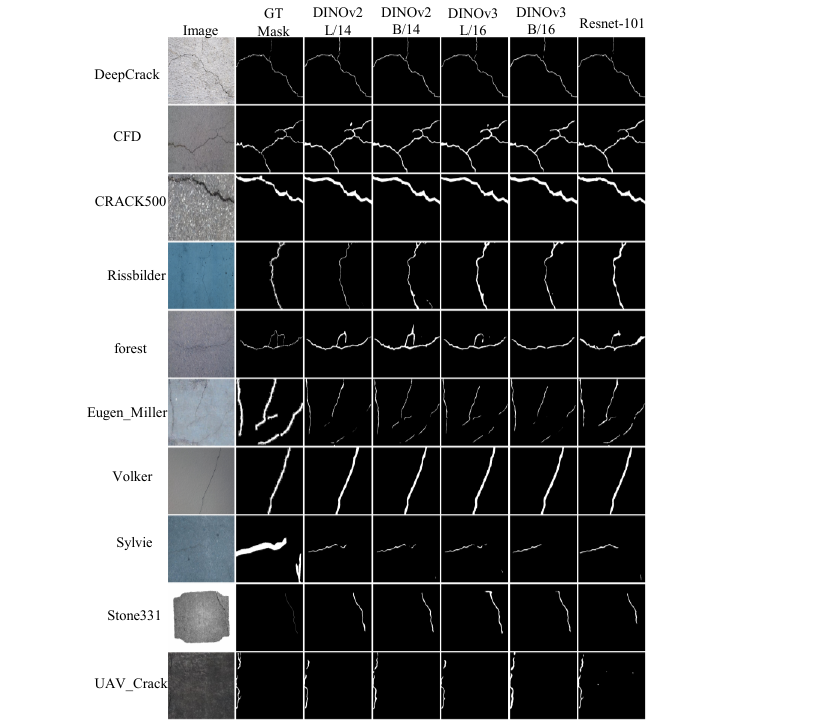}
    \caption{CrackGeoFM predictions with five backbone architectures 
    across in-domain and zero-shot datasets.}
    \label{fig:backbone_predictions}
\end{figure}

Figure~\ref{fig:backbone_predictions} illustrates these 
differences visually. The four ViT backbones produce 
nearly indistinguishable predictions on in-domain datasets 
(DeepCrack, CFD, CRACK500), consistent with the narrow 
1.2\% Dice range reported in Table~\ref{tab:backbone_ablation}. 
On zero-shot datasets, ResNet-101 produces visibly incomplete 
or noisier predictions (Rissbilder, Eugen\_Miller, Stone331), 
confirming that the 11\% zero-shot Dice gap reflects 
genuinely weaker transfer rather than a metric artifact. 
The ViT backbones' self-supervised patch-level representations 
provide more robust spatial features for unseen crack domains 
that the CFAM adapters can effectively specialize.

\subsection{Statistical Analysis}
Statistical significance is assessed using the Friedman test for multi-model comparison and the Wilcoxon signed-rank test for pairwise comparisons, with Benjamini-Hochberg (BH) correction for multiple comparisons at $\alpha = 0.05$. Effect sizes are reported as Cohen's $d$. Because comparisons are conducted across datasets (7 val, 6 zero-shot), sample sizes are small, which limits statistical power particularly on zero-shot evaluation.

\begin{figure}[H]
    \centering
    \includegraphics[width=\columnwidth]{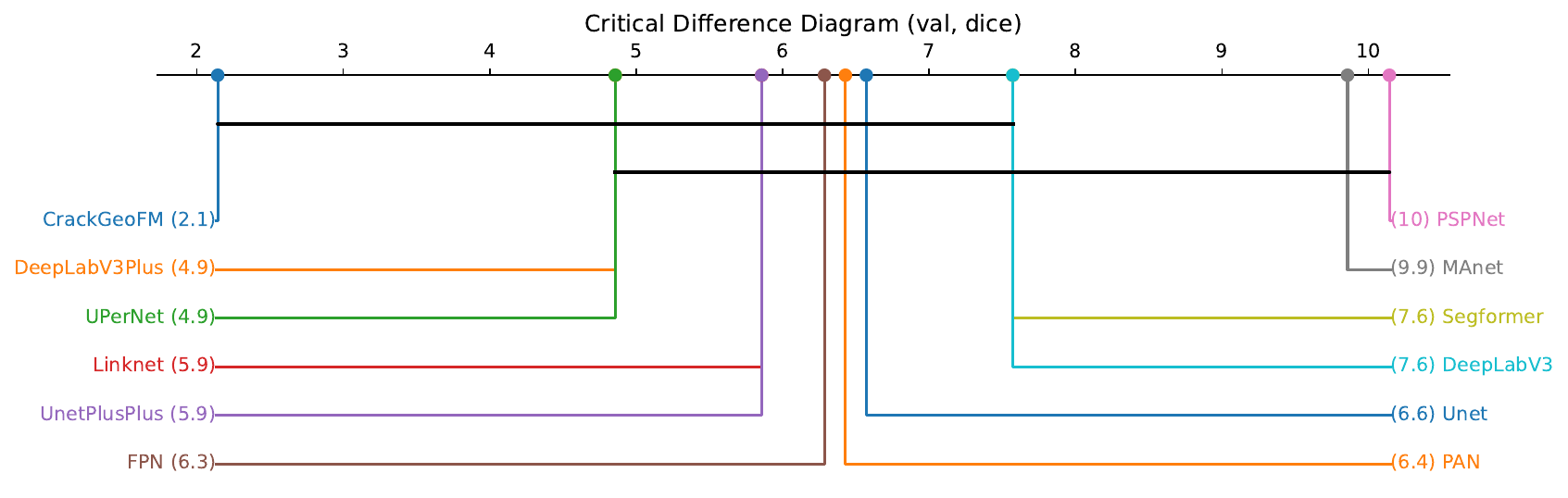}
    \caption{Critical difference diagram for Dice on validation 
    datasets (Friedman test, $p = 0.003$). Models are ranked by 
    average performance; lower rank is better. Connected models 
    (joined by a horizontal bar) are not significantly different 
    by the Nemenyi post-hoc test ($\alpha = 0.05$). CrackGeoFM 
    achieves the best average rank (2.1).}
    \label{fig:cd_diagram}
\end{figure}

On validation data, the Friedman test confirms significant differences among the 12 models for both Dice ($\chi^2 = 28.05$, $p = 0.003$) and MCC ($\chi^2 = 29.79$, $p = 0.002$). CrackGeoFM achieves the best average rank (2.1) across all validation datasets, as shown in the critical difference diagram (Figure~\ref{fig:cd_diagram}). Pairwise Wilcoxon tests indicate that CrackGeoFM significantly outperforms 8 of 11 baselines after BH correction ($p < 0.05$), with large effect sizes (Cohen's $d > 0.7$) against PSPNet, MAnet, DeepLabV3, DeepLabV3+, FPN, SegFormer, PAN, and LinkNet. The three non-significant comparisons are against U-Net ($p = 0.074$, $d = 0.90$), UPerNet ($p = 0.057$, $d = 1.28$), and U-Net++ ($p = 0.375$, $d = 0.38$), where the differences are consistent in direction (CrackGeoFM wins on 5-7 of 7 datasets) but do not reach statistical significance at the corrected threshold.

On zero-shot data, the Friedman test does not reach significance ($\chi^2 = 16.15$, $p = 0.136$), and no pairwise comparison survives BH correction. This is expected: with only six held-out datasets, the Wilcoxon test has limited power to detect differences at $\alpha = 0.05$. However, CrackGeoFM consistently ranks first (winning 4 of 6 datasets against every baseline), and the effect sizes, while classified as small to medium (Cohen's $d = 0.20$-$0.52$), are uniformly positive. The lack of significance reflects insufficient sample size rather than absence of improvement, as confirmed by the per-dataset results in Section~\ref{sec:zeroshot} where CrackGeoFM achieves the largest absolute gains on the most challenging transfer cases.

\begin{figure}[H]
    \centering
    \includegraphics[width=\columnwidth]{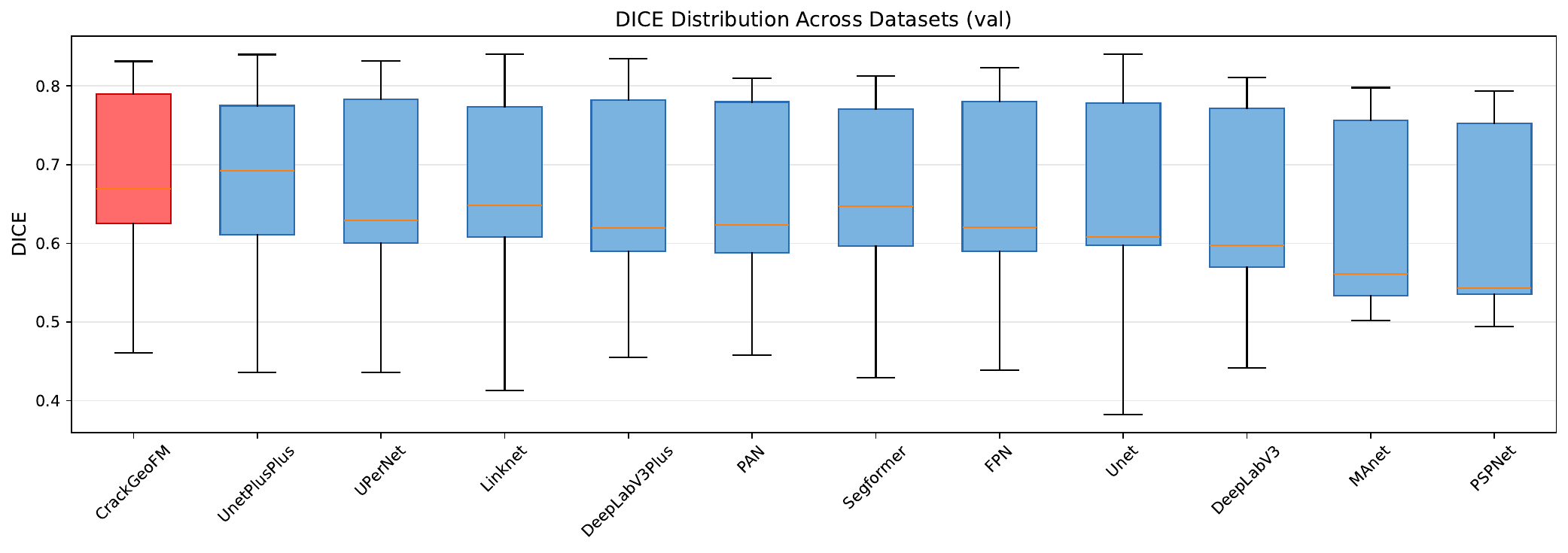}
    \vspace{0.5em}
    \includegraphics[width=\columnwidth]{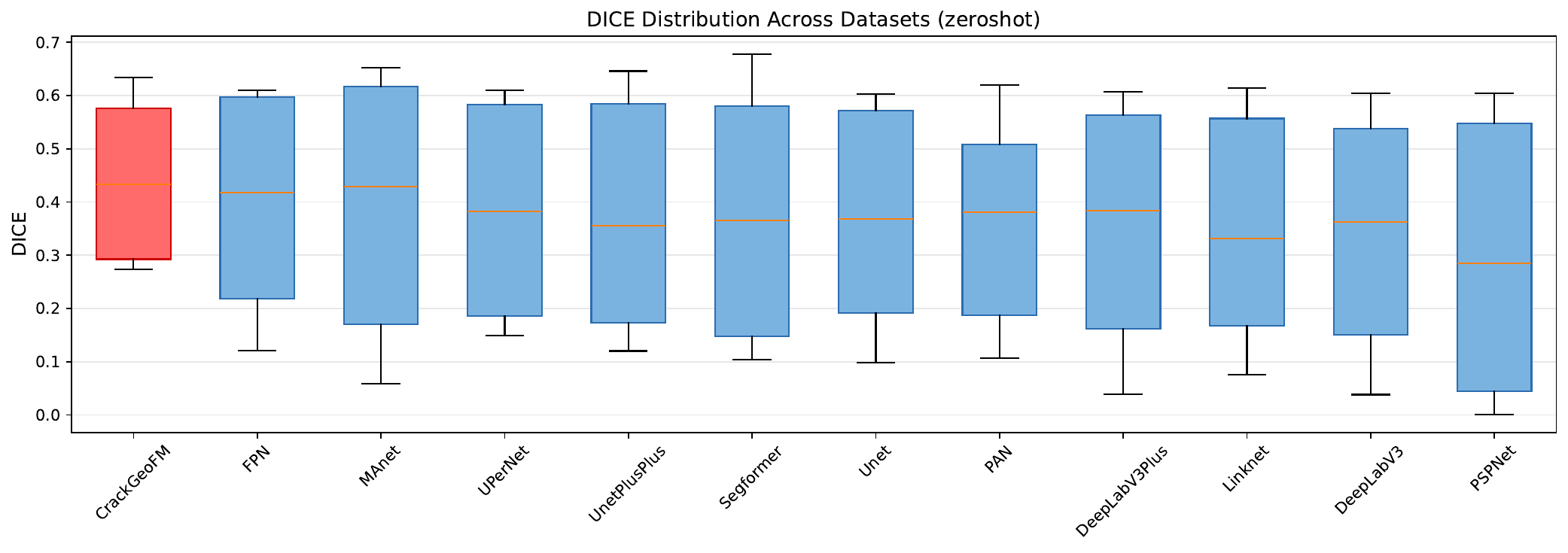}
    \caption{Dice distributions across validation (top) and 
    zero-shot (bottom) datasets. CrackGeoFM (red) achieves the 
    highest median on both tiers with a narrower interquartile 
    range on zero-shot data, indicating more consistent 
    cross-domain performance.}
    \label{fig:boxplot}
\end{figure}

Figure~\ref{fig:boxplot} presents the Dice distributions across datasets. On validation data, CrackGeoFM exhibits the highest median and a comparable interquartile range to the top baselines. On zero-shot data, CrackGeoFM maintains the highest median with a narrower spread, indicating more consistent cross-domain performance.

\subsection{Computational Complexity}
Table~\ref{tab:complexity} compares the computational cost of CrackGeoFM against all baselines in terms of total parameters, trainable parameters, floating-point operations, inference latency, and peak GPU memory, measured on a single NVIDIA A40 GPU with 448$\times$448 input resolution.

CrackGeoFM has a substantially larger parameter count (310.9M) than any baseline (24.3-147.4M) because the frozen DINOv2-Large backbone accounts for 304.4M parameters (97.9\% of the total). However, only 6.5M parameters (2.1\%) are trainable, which is smaller than every baseline where all parameters are updated during training. This asymmetry is a deliberate design choice: the frozen backbone preserves the pretrained representations that enable cross-domain transfer, while the small trainable footprint reduces overfitting risk and enables rapid few-shot adaptation with as few as five labeled images (Section~\ref{sec:fewshot}).

The computational cost follows a similar pattern. CrackGeoFM requires 347.6 GFLOPs and 104.3 ms per image (9.6 FPS), compared to 9.0-176.2 GFLOPs and 3.1-25.8 ms (38.8-317.7 FPS) for the baselines. The higher cost arises primarily from the 24-block ViT-L forward pass and the multi-scale FPN aggregation of four adapter outputs. Peak GPU memory is 1,272 MB, roughly 1.5-5$\times$ that of most baselines, though well within the capacity of standard inference hardware.

Two factors contextualize this overhead. First, CrackGeoFM produces three outputs (mask, skeleton, uncertainty) in a single forward pass, whereas baselines produce only the mask and would require separate models or post-processing steps to obtain equivalent structural and confidence information. Second, infrastructure crack inspection is predominantly an offline workflow where images are collected during surveys and processed in batch. In this setting, throughput of 9.6 FPS is sufficient for practical deployment, and the improved generalization and multi-output capability justify the additional compute relative to single-output baselines operating at higher frame rates.

\begin{table}[t]
\centering
\caption{Computational complexity comparison. All measurements use 448$\times$448 input on a single NVIDIA A40 GPU.}
\label{tab:complexity}
\resizebox{\columnwidth}{!}{%
\begin{tabular}{lccccccc}
\toprule
\textbf{Model} & \textbf{Params (M)} & \textbf{Train. (M)} & \textbf{GFLOPs} & \textbf{Latency (ms)} & \textbf{FPS} & \textbf{Mem (MB)} & \textbf{Outputs} \\
\midrule
PSPNet          & 24.3  & 24.3  & 9.0   & 3.1  & 317.7 & 160  & 1 \\
PAN             & 24.3  & 24.3  & 26.6  & 7.1  & 140.1 & 168  & 1 \\
SegFormer       & 24.8  & 24.8  & 23.0  & 8.0  & 124.8 & 340  & 1 \\
FPN             & 26.1  & 26.1  & 24.0  & 6.7  & 148.8 & 218  & 1 \\
DeepLabV3+      & 26.7  & 26.7  & 28.1  & 7.1  & 140.2 & 210  & 1 \\
LinkNet         & 31.2  & 31.2  & 22.4  & 7.7  & 129.6 & 240  & 1 \\
U-Net           & 32.5  & 32.5  & 32.7  & 8.1  & 124.1 & 240  & 1 \\
UPerNet         & 37.3  & 37.3  & 58.7  & 10.0 & 100.4 & 425  & 1 \\
DeepLabV3       & 39.6  & 39.6  & 125.4 & 15.6 & 64.1  & 289  & 1 \\
U-Net++         & 49.0  & 49.0  & 176.2 & 25.8 & 38.8  & 838  & 1 \\
MAnet           & 147.4 & 147.4 & 57.1  & 13.1 & 76.5  & 765  & 1 \\
\midrule
CrackGeoFM      & 310.9 & \textbf{6.5} & 347.6 & 104.3 & 9.6 & 1,272 & 3 \\
\bottomrule
\end{tabular}%
}
\end{table}

\subsection{Practical Implications}
CrackGeoFM's multi-output design has direct implications for how crack inspection workflows are organized in practice. Conventional pipelines require separate models or post-processing stages to obtain segmentation masks, structural measurements, and confidence estimates, each introducing potential inconsistencies and additional computational overhead. CrackGeoFM produces all three outputs in a single forward pass, ensuring that the skeleton is geometrically consistent with the mask and the uncertainty map reflects the actual prediction confidence, rather than being derived independently after the fact.

The few-shot adaptation capability addresses a persistent barrier to deploying learned models in infrastructure management. Agencies responsible for bridges, tunnels, dams, or building facades typically lack the resources to curate large labeled datasets for each new structure type. CrackGeoFM enables deployment to a new inspection context by labeling as few as five representative images from the target domain and fine-tuning only the lightweight adapters, a process that takes minutes on a single GPU and requires no machine learning expertise beyond basic script execution. The frozen backbone ensures that the adapted model retains its general crack knowledge while specializing for the local domain.

The calibrated uncertainty output supports a risk-informed review strategy that reduces human workload without compromising inspection coverage. Rather than manually reviewing all predictions, inspectors can focus on regions flagged as high-uncertainty, where the model's ECE below 0.02 ensures that confidence scores reliably indicate prediction quality. This selective review is particularly valuable for large-scale surveys where thousands of images must be processed per inspection cycle, allowing human attention to be directed where it is most needed.

The skeleton output enables automated crack length and connectivity measurements that are difficult to extract reliably from binary masks alone. Because the skeleton preserves branching structure and geometric continuity, it can support downstream engineering tasks such as crack growth monitoring across inspection cycles, severity classification based on extent and connectivity, and prioritization of maintenance interventions based on spatial crack patterns rather than aggregate pixel counts.

\section{Limitations and Future Work}
Despite the strong results across multiple evaluation tiers, several limitations warrant discussion and suggest directions for future research.

The most significant computational limitation is inference speed. CrackGeoFM processes images at 9.6 FPS due to the 24-block ViT-L backbone, which is approximately 10--30$\times$ slower than the lightweight baselines. While this throughput is adequate for offline batch processing of survey images, it precludes real-time deployment on edge devices or resource-constrained inspection platforms. Future work could address this through knowledge distillation into a smaller student network that retains the multi-task outputs and cross-domain transfer capability while reducing the backbone to a ViT-Small or efficient CNN architecture.

The zero-shot evaluation on Eugen\_Miller reveals that the model defaults to a conservative failure mode when the target domain differs substantially from all training data. The frozen backbone features lack sufficient crack evidence for tunnel imagery, and the uncertainty-aware loss suppresses predictions rather than producing false positives. While few-shot adaptation resolves this limitation with as few as five labeled images, a fully zero-shot solution would require either broader training data coverage or domain-adaptive mechanisms that can adjust backbone features at inference time without labeled target data.

The current skeleton targets are derived through morphological processing of ground truth masks, which introduces approximation errors from the closing, thinning, and pruning pipeline. These targets are sufficient for learning general crack topology, but they do not capture precise geometric properties such as crack width profiles or curvature that engineering assessments often require. Future work could incorporate explicit crack width estimation as a fourth output head, extending the multi-task framework toward more complete quantitative crack characterization.

The uncertainty estimation captures aleatoric uncertainty through learned per-pixel variance but does not model epistemic uncertainty arising from limited training data. Incorporating epistemic uncertainty through Monte Carlo dropout or deep ensembles would provide a more complete confidence picture, particularly for out-of-distribution detection where the model encounters infrastructure types entirely absent from the training pool.

Finally, the current framework operates on individual images without exploiting temporal or spatial relationships between consecutive inspection frames. Extending CrackGeoFM to video-based or sequential inspection settings could enable crack growth monitoring over time, utilizing the skeleton representation for tracking geometric changes across inspection cycles.

\section{Conclusion}
This paper presented CrackGeoFM, a multi-task crack-domain foundation model that jointly produces segmentation masks, crack skeletons, and per-pixel uncertainty from a single forward pass. The model augments a frozen DINOv2-Large backbone with three modules: FCEM for wavelet-derived high-frequency crack cues, CFAM for lightweight domain adaptation, and SMTD for multi-scale multi-task decoding. All skeleton targets are derived automatically from standard binary masks, requiring no additional annotation. The effectiveness of this design is validated across 20 crack datasets spanning five surface materials and four imaging platforms.

CrackGeoFM achieves the highest mean Dice on both in-domain (0.6843) and zero-shot evaluation (0.4406, +4.5\%), with the most striking gain on LCSD (+37.8\%), where frequency-enhanced features maintain crack sensitivity under low-light conditions that cause baseline models to fail. The model exhibits the smallest generalization gap (0.244) among all compared methods, and its direct skeleton prediction outperforms post-hoc skeletonization by +4.9\% SkelDice. Uncertainty calibration transfers reliably to unseen domains (ECE below 0.02), and few-shot adaptation with five labeled images recovers performance on the most challenging held-out datasets, with Eugen\_Miller improving by +38.8\% at $k{=}5$ and +41.8\% at $k{=}25$. Understanding these gains requires examining how the modules interact.

The ablation study reveals that the modules are complementary rather than independently additive: skeleton supervision and frequency features reduce performance in isolation, but the topology consistency loss aligns the competing objectives and enables the full architecture to surpass all intermediate configurations. These findings establish CrackGeoFM as an integrated system whose components reinforce each other.

Providing mask, skeleton, and calibrated uncertainty in a unified framework enables CrackGeoFM to support the full spectrum of engineering crack analysis, from pixel-level detection through geometric measurement to confidence-aware decision support.

\section{Data Availability}
All data used for this research are publicly available. 

\section{Acknowledgements}
This research was supported by the United States Department of Transportation (USDOT) under federal grant number 69A3552348308, administered at the institutional level under grant number FAR0038979. The contents of this paper reflect the views of the authors, who are solely responsible for the facts and accuracy of the information presented herein. The contents do not necessarily reflect the official views or policies of the United States Department of Transportation.

\clearpage
\bibliographystyle{elsarticle-num}
\bibliography{export}
\end{document}